\documentclass[singlecolumn,10pt,twoside,final]{asme2ej_Corrected}
\usepackage{amsfonts}
\usepackage{graphicx}
\usepackage{amsfonts}
\usepackage{graphicx}
\usepackage{amsmath, amssymb, bm}
\usepackage[fixamsmath,disallowspaces]{mathtools}
\usepackage{xcolor}
\usepackage{fixmath}
\usepackage{soul}

\newtheorem{remark}{Remark}

\input{tcilatex}
\begin{document}

\parindent 0pt \parskip 2pt \setcounter{topnumber}{9} %
\setcounter{bottomnumber}{9} \renewcommand{\textfraction}{0.00001}

\renewcommand {\floatpagefraction}{0.999} \renewcommand{\textfraction}{0.01} %
\renewcommand{\topfraction}{0.999} \renewcommand{\bottomfraction}{0.99} %
\renewcommand{\floatpagefraction}{0.99} \setcounter{totalnumber}{9}

\input{opening_HamelCoefficients_Corrected.inp}

\section{Introduction}

The Boltzmann-Hamel (BH) equations are a milestone result of the research on
the dynamics of nonholonomic systems, which was initiated by Hertz \cite%
{Hertz1894}. Hamel continued previous works of Volterra \cite{Volterra1898}
(who continued the work of Beltrami), Tz\'{e}noff \cite%
{Tzenoff1920,Tzenoff1924} (who built upon the work of Chaplygin \cite%
{Chaplygin1897a,Chaplygin1897b,Chaplygin1911}), Poincar\'{e} \cite%
{Poincare1892}, Voronets \cite{Voronets1901,Woronetz1910}, and Boltzmann 
\cite{Boltzmann1902} in particular, into a general concept. Key elements in
these equations are the three-index symbols that are referred to as 'Hamel
coefficients', which reveal the nonholonomic nature of the quasi-coordinates
respectively of the constraints a system is subjected to.

If the Lagrangian is invariant under the action of a Lie group, then the
Hamel coefficients are identical to the structure constants of this group.
The BH-equations are then the Euler-Lagrange equations corresponding to a
reduced variational principle on this symmetry group, respectively the
Euler-Poincar\'{e} equations \cite%
{BlochMarsdenZenkov2009,MarsdenBook,MarsdenScheurle1993}. For this reason
the BH-equations are often used as synonym for the reduced Euler-Lagrange
equations \cite{BallZenkov2015,BlochMarsdenZenkov2009}. The configuration
space of a rigid body is the Lie group $SE\left( 3\right) $ (the special
Euclidean group \cite{MarsdenScheurle1993,MarsdenBook,Murray,Selig}), and
the BH-equations are the Euler-Poincar\'{e} equations (when formulated in
terms of body-fixed twists), or the reduced Euler-Lagrange equations (when
using spatial twists). The rigid body is often used as the typical example
to study fundamental aspects of nonholonomic systems in a coordinate-free
geometric setting. However, a systematic treatment of the original
BH-equations \cite{Hamel1904,Hamel-MathAnal1924,HamelBook} in local
coordinates along with the original derivation of the Hamel coefficients for
the rigid body, in a way that links them to the Euler-Lagrange equations on $%
SE\left( 3\right) $, has not yet been reported in the literature. Key to
this is evaluating the Hamel coefficients taking into account the geometry
of rigid body motions (possibly without resorting to local coordinates),
which not only demonstrates technicalities of the actual determination of
these coefficients, but also offers insight into the geometric nature of the
problem. Moreover, the Hamel coefficients are specific to the particular
representation of rigid body velocities (i.e. the definition of
quasi-velocities) and the corresponding invariance of the rigid body
Lagrangian under action of the symmetry group. The three most relevant
representations of rigid body velocities, are the body-fixed, spatial, and
hybrid representation \cite{MUBOScrews2,Murray}. The first two correspond to
a left-invariant and right-invariant Lagrangian on $SE\left( 3\right) $,
respectively. The hybrid representation assumes the direct product group $%
SO\left( 3\right) \times {\mathbb{R}}^{3}$ as symmetry group and
corresponding right-invariant Lagrangian, which clearly implies a different
geometric model of rigid body motions \cite{BIT2016,MMT2014}. The
derivations of the Hamel coefficients for the rigid body reported in the
literature almost exclusively use particular local coordinates to describe
rigid body motions (see e.g. \cite[pp. 18-20]{Hamel1904} or \cite[p. 369]%
{PapastavridisBook2002} for the rotating body, and \cite[pp. 43 \& 217]%
{Whittaker1904} or \cite[p. 76]{LurieBook2002} for the free body), which
leads to rather involved expressions. Some of the few exception are Hamel's
derivation for the rotating body described in terms of the body-fixed
angular velocity (see \cite[p. 489]{HamelBook}), and the derivation for the
rotating $n$-dimensional body by Hurtado \& Sinclair \cite{Hurtado2004}.

In this paper, a systematic derivation of the Hamel coefficients for the
rigid body and the corresponding BH-equations are presented for the three
representations of rigid body velocities mentioned above. The Hamel
coefficients are derived without resorting to local coordinates (no
parameterization of rigid body motions). Hamel's Euler-Lagrange equations
equally apply to systems subjected to Pfaffian constraints, where the
nonholonomic geometry is encoded in the Hamel coefficients. This is recalled
and several formulations of the Hamel equations are presented. The Hamel
equations for the rolling ball with inhomogenous mass density are presented,
which is a typical example for a nonholonomically constrained rigid body 
\cite{HolmBook2}.

Despite the long history of analytical rigid body dynamics, this paper seems
to be the first publication reporting an explicit geometric construction of
the Hamel coefficients, which are indeed determined by the structure
constant of the Lie group underlying the respective definition of rigid body
twist. The paper summarizes material that is scattered in the literature
along with references to the original sources, and presents novel
derivations of Hamel coefficients. The paper is organized as follows.
Section \ref{secEOM} briefly summarizes the BH-equations in their original
form and the definition of the Hamel coefficients. First the BH-equations
for general unconstrained systems described in quasi-coordinates are
reviewed. Then the Hamel equations for constrained systems described in
quasi-coordinates as well as in 'holonomic' coordinates are recalled. In
order to make a connection to the modern approach in geometric mechanics,
the BH-equations for the rigid body are finally identified as the
Euler-Poincar\'{e} respectively the reduced Euler-Lagrange equations when
using $SE\left( 3\right) $ as rigid body configuration space. The Hamel
coefficients are derived in section \ref{secHamelCoeff} for the rotating
rigid body when the angular velocity in body-fixed and spatial
representation is used as quasi-velocity. The corresponding BH-equations are
derived in section \ref{secBHERot}. The Hamel coefficients for the free
rigid body are derived in section \ref{secHamelCoeffFree} where the
body-fixed, spatial, and hybrid representations of twists are used as
quasi-velocities. The corresponding BH-equations are derived explicitly in
section \ref{secBHEFree}. In addition, the Hamel equations for an
inhomogenous ball rolling on a horizontal plane are derived as an example
for a nonholonomically constrained rigid body. A list of symbols and
abbreviations is given in appendix \ref{secSymbols} and \ref{secAbbr},
respectively. Throughout the paper, the Lagrangian will be the kinetic
energy.

Notation: With reference to the original development by Hamel and Boltzmann,
the classical notation in terms of local coordinates is used. Tensor
notation and the Ricci summation convention is applied throughout the paper.
Cartesian vectors are resolved in orthogonal frames, and their covariant
coordinates (lower indices) are used. $\varepsilon _{ijk}$ denotes the
Levi-Civita symbol, and $\delta _{ij}$ the Kronecker symbol.

\section{Motion Equations in Quasi-Velocities%
\label{secEOM}%
}

\subsection{Generalized coordinates and quasi-velocities}

The configuration of the system is described by local coordinates $%
q^{a},a=1,\ldots ,n$ summarized in the vector $\mathbf{q}\in {\mathbb{V}}%
^{n} $. The system velocity is described by the (quasi)velocity coordinates $%
u^{a},a=1,\ldots ,n$, summarized in $\mathbf{u}\in {\mathbb{R}}^{n}$, which
are related to the generalized speeds by%
\begin{equation}
\mathbf{u}=\mathbf{A}\left( \mathbf{q}\right) \dot{\mathbf{q}}  \label{u2}
\end{equation}%
and in components%
\begin{equation}
u^{a}=A_{b}^{a}\dot{q}^{b}.  \label{u}
\end{equation}%
The transition to the corresponding (quasi)coordinates $\pi ^{a},a=1,\ldots
,n$ is described by the Pfaffian forms (not to be confused with the angular
velocity)%
\begin{equation}
\omega ^{a}:=d\pi ^{a}=A_{b}^{a}dq^{b}.  \label{dpi}
\end{equation}%
If the relation (\ref{dpi}) is not integrable for some $a$, then the $\pi
^{a}$ are referred to as quasi-coordinates, and the $u^{a}$ as
quasi-velocities (also called 'nonholonomic' velocities, see remark \ref%
{remHolonCoord}).

The inverse relation to (\ref{u2}) is written as%
\begin{equation}
\dot{\mathbf{q}}=\mathbf{B\left( \mathbf{q}\right) u}  \label{u3}
\end{equation}%
with $\mathbf{B}:=\mathbf{A}^{-1}$, and in components%
\begin{equation}
\dot{q}^{a}=B_{b}^{a}u^{b}.  \label{qdu}
\end{equation}%
Accordingly, the inverse of (\ref{dpi}) is $dq^{a}=B_{b}^{a}d\pi ^{b}$. In
the following, the transformations are assumed to be regular.

For a given trajectory $q^{a}\left( t\right) $, the (quasi)velocities can be
integrated to obtain (quasi)coordinates%
\begin{equation}
\pi ^{a}:=\int u^{a}dt=\int A_{b}^{a}\dot{q}^{b}dt=\int \omega ^{a}=\int
A_{b}^{a}dq_{b}.
\end{equation}%
The so obtained quantities can serve as generalized coordinates if and only
if the last integral on the right-hand side is path independent. This is so
if and only if $d\omega ^{a}\equiv 0$ (i.e. $\omega ^{a}$ are exact
one-forms), and in coordinates%
\begin{equation}
\frac{\partial A_{b}^{a}}{\partial q^{c}}-\frac{\partial A_{c}^{a}}{\partial
q^{b}}\equiv 0  \label{IntCond}
\end{equation}%
which are the conditions for integrability of the system of the $n$ Pfaffian
forms (\ref{dpi}).

\subsection{Rigid body velocities}

Rigid body motions evolve on the Lie group $SE\left( 3\right) $, and the
quasi-velocity is the body twist. In body-fixed representation, the latter
is given by the left invariant-invariant vector field (\ref{BodyFixedTwist})
on $SE\left( 3\right) $. The motion of a rigid body rotating about a fixed
point evolves on $SO\left( 3\right) $, and the quasi-velocity is the angular
velocity. In body-fixed representation this is the left-invariant vector
field (\ref{w1}) on $SO\left( 3\right) $. Thus, (\ref{dpi}) are the
Maurer-Cartan forms on $SE\left( 3\right) $ and $SO\left( 3\right) $,
respectively. They satisfy the Maurer-Cartan equations \cite[p. 255]%
{MarsdenBook} 
\begin{equation}
d\omega ^{a}=-\frac{1}{2}c_{bc}^{a}\omega ^{b}\wedge \omega ^{c}  \label{dw}
\end{equation}%
which are in local coordinates%
\begin{equation}
\frac{\partial A_{b}^{a}}{\partial q^{c}}-\frac{\partial A_{c}^{a}}{\partial
q^{b}}=c_{rs}^{a}A_{b}^{r}A_{c}^{s}  \label{MC}
\end{equation}%
where $c_{bc}^{a}$ are the structure constants of the Lie group. Clearly,
the condition (\ref{IntCond}) is not satisfied for any non-Abelian group.
This holds true for $SE\left( 3\right) $ and $SO\left( 3\right) $, asserting
that rigid body velocities are non-integrable. The same argument applies to
right-invariant vector fields, i.e. when using spatial representations of
rigid body velocities, where the sign in (\ref{dw}) changes. For the Abelian
motion subgroups ${\mathbb{R}}^{n},SO\left( 2\right) ,SO\left( 2\right)
\times {\mathbb{R}}$, the velocities are indeed integrable.

\subsection{The Boltzmann-Hamel equations}

Let $L\left( q^{a},u^{a}\right) $ be the Lagrangian in terms of
(quasi)velocity coordinates $u^{a}$. For mechanical systems this is usually $%
L\left( q^{a},u^{a}\right) =T\left( q^{a},u^{a}\right) -U\left( q^{a}\right) 
$ with the kinetic (co)energy $T\left( q^{a},u^{a}\right) $ and the
potential energy $U\left( q^{a}\right) $, but can be a general energetic
state function for a general dynamic system, e.g. given by the magnetic
coenergy and electric energy for discrete electromechanical systems \cite%
{MaisserEMS}. The variational form of the Hamel equations is \cite%
{Hamel1904,Hamel-MathAnal1924,HamelBook,PapastavridisBook2002}%
\begin{equation}
\left( \frac{d}{dt}\frac{\partial L}{\partial u^{a}}-\frac{\partial L}{%
\partial \pi ^{a}}-Q_{a}\right) \delta \pi ^{a}+\frac{\partial L}{\partial
u^{a}}\left( \frac{d\delta \pi ^{a}-\delta d\pi ^{a}}{dt}\right) =0
\label{BHLagrange}
\end{equation}%
where $Q_{a}$ are generalized forces dual to $u^{a}$. The term in the second
bracket satisfies the \emph{transitory (or transpositional) relation} 
\begin{equation}
\frac{d\delta \pi ^{a}-\delta d\pi ^{a}}{dt}=\gamma _{cb}^{a}u^{b}\delta \pi
^{c}  \label{transpos}
\end{equation}%
respectively $d\delta \pi ^{a}-\delta d\pi ^{a}=\gamma _{cb}^{a}d\pi
^{c}\delta \pi ^{c}$, where, with the transformation (\ref{u}),%
\begin{equation}
\gamma _{ab}^{c}:=\left( \frac{\partial A_{r}^{c}}{\partial q^{s}}-\frac{%
\partial A_{s}^{c}}{\partial q^{r}}\right) B_{a}^{r}B_{b}^{s}  \label{gamma}
\end{equation}%
are the \emph{Hamel coefficients}. The \emph{Boltzmann-Hamel equations }in
their explicit form are hence \cite%
{Hamel1904,Hamel-MathAnal1924,HamelBook,NeimarkFufaev1972,PapastavridisBook2002}%
\begin{equation}
\frac{d}{dt}\frac{\partial L}{\partial u^{a}}-\frac{\partial L}{\partial \pi
^{a}}+\frac{\partial L}{\partial u^{b}}\gamma _{ac}^{b}u^{c}=Q_{a}.
\label{BHEqu}
\end{equation}%
Complemented with (\ref{qdu}) they describe the dynamics of an unconstrained
system in terms of (quasi)coordinates.

Since the Lagrangian depends on the coordinates $q^{a}$, rather than on the
quasi-coordinates $\pi ^{a}$, for the actual evaluation of (\ref{BHEqu}) the
partial derivative $\frac{\partial L}{\partial \pi ^{a}}$ is replaced using
the Helmholtz identity $\frac{\partial q^{b}}{\partial \pi ^{a}}=\frac{%
\partial \dot{q}^{b}}{\partial u^{a}}=B_{b}^{a}$. This yields the
'practical' form of the Hamel equations%
\begin{equation}
\frac{d}{dt}\frac{\partial L}{\partial u^{a}}-\frac{\partial L}{\partial
q^{b}}B_{a}^{b}+\frac{\partial L}{\partial u^{b}}\gamma _{ac}^{b}u^{c}=Q_{a}.
\label{BHEqu2}
\end{equation}%
Comparing (\ref{gamma}) and (\ref{IntCond}) shows that the Hamel
coefficients vanish if and only if $u^{a}$ are integrable.

Now consider a system defined on a Lie group. With regular $\mathbf{A}$,
relation (\ref{MC}) is transformed to%
\begin{equation}
c_{ab}^{c}=\left( \frac{\partial A_{r}^{c}}{\partial q^{s}}-\frac{\partial
A_{s}^{c}}{\partial q^{r}}\right) B_{a}^{r}B_{b}^{s}.
\end{equation}%
Comparing this with (\ref{gamma}) reveals that the Hamel coefficients are
identical to the structure constants (with negative sign if spatial
representation is used). This already shows that the Hamel coefficients for
the rigid body are the structure constants of $SE\left( 3\right) $,
respectively of $SO\left( 3\right) \times {\mathbb{R}}^{3}$. The goal of
this paper is to derive them from their original definition (\ref{gamma}).

\begin{remark}
The equations (\ref{BHEqu}) were published by Hamel in \cite%
{Hamel1904,Hamel-MathAnal1924,HamelBook} who referred to them as the
Lagrange-Euler equations. Because Boltzmann \cite%
{Boltzmann1902,Boltzmann1904} presented a version of Lagrange's equations
which is very close to Hamel's equations, (\ref{BHEqu}) are usually referred
to as the Boltzmann-Hamel equations \cite{Maisser1997,MaruskinBloch2007}.
Further, Lurie \cite{LurieBook2002} called $\gamma _{ab}^{c}$ the Boltzmann
'three-index symbols'. In the context of parametrization of rotations, they
are called Ricci coefficients.
\end{remark}

\begin{remark}
The expression (\ref{gamma}) can be reformulated as%
\begin{eqnarray}
\gamma _{ab}^{c} &=&\left( B_{a}^{r}B_{b}^{s}-B_{a}^{s}B_{b}^{r}\right) 
\frac{\partial A_{r}^{c}}{\partial q^{s}}  \label{gamma2} \\
&=&A_{r}^{c}\left( \frac{\partial B_{b}^{r}}{\partial q^{a}}-\frac{\partial
B_{a}^{r}}{\partial q^{b}}\right) .  \label{gamma3}
\end{eqnarray}%
With regularity of $\mathbf{A}$, the last form gives rise to 
\begin{equation}
\frac{\partial B_{b}^{r}}{\partial q^{a}}-\frac{\partial B_{a}^{r}}{\partial
q^{b}}=\gamma _{ab}^{c}B_{c}^{r}.
\end{equation}%
These expressions may be advantageous for actual evaluations.
\end{remark}

\begin{remark}
If $u^{a}$ are integrable, then $u^{a}=\dot{q}^{a}$ (or linear combinations
of $\dot{q}^{a}$), and (\ref{BHEqu}) reduce to the ordinary Euler-Lagrange
equations on $T{\mathbb{V}}^{n}$%
\begin{equation}
\frac{d}{dt}\frac{\partial L}{\partial \dot{q}^{a}}-\frac{\partial L}{%
\partial q^{b}}=Q_{a}  \label{LE}
\end{equation}%
and the $q^{a}$ are called Lagrange coordinates.
\end{remark}

\subsection{Hamel equations for constrained systems}

For completeness, and since it is instructive to trace back the origin of
the terminology, the \emph{Hamel equations} for a system described in terms
of quasi-velocities subjected to Pfaffian constraints are recalled. The
naming shall reflect the fact that these equations were introduced by Hamel
whereas Boltzmann exclusively considered unconstrained systems.

\subsubsection{Hamel-Equations in terms of quasi-velocities%
\label{secBHContrained}%
}

The point of departure is the Lagrangian $L\left( q^{a},\dot{q}^{a}\right) $
in terms of 'holonomic' coordinates. As before, $q^{a},1,\ldots ,n$ are
coordinates used to describe the configuration of the system, but now the $%
\dot{q}^{a}$ are subjected to a system of $\bar{m}$ Pfaffian constraints%
\begin{equation}
u^{\alpha }=0\text{ \ \ with \ }u^{\alpha }:=A_{a}^{\alpha }\dot{q}^{a},\ \
\alpha =1,\ldots ,\bar{m}  \label{PfaffianConstr}
\end{equation}%
so that only $\delta :=n-\bar{m}$ of the $\dot{q}^{a}$ are independent (for
simplicity catastatic constraints are considered). Thus $n-\bar{m}$
independent quasi-velocities can be introduced as%
\begin{equation}
u^{I}=A_{a}^{I}\dot{q}^{a},\ \ I=\bar{m}+1,\ldots ,n.  \label{PfaffForm}
\end{equation}%
These independent velocity coordinates are indexed with capital Lattin
letters $I,J,K,\ldots =\bar{m}+1,\ldots ,n$ (w.l.o.g. the last $n-\bar{m}$
indices). The system of Pfaffian equations (\ref{PfaffianConstr}) and (\ref%
{PfaffForm}) are summarized as in (\ref{u}). The solution of this system is
given by (\ref{qdu}), presuming that $\mathbf{A}$ is regular.

Denote with $\mathcal{L}\left( q^{a},u^{a}\right) :=L\left(
q^{a},B_{b}^{a}u^{b}\right) $ the Lagrangian after replacing $\dot{q}^{a}$
by $u^{a}$ according to (\ref{qdu}). The Hamel equations of the constrained
system in terms of the (quasi)velocities $u^{I}$ are \cite%
{Hamel1904,Hamel-MathAnal1924,HamelBook}%
\begin{equation}
\frac{d}{dt}\frac{\partial \mathcal{L}}{\partial u^{I}}-\frac{\partial 
\mathcal{L}}{\partial q^{a}}B_{I}^{a}+\frac{\partial \mathcal{L}}{\partial
u^{a}}\gamma _{IJ}^{a}%
\red%
u^{J}%
\color{black}%
=Q_{I},\ \ I=\bar{m}+1,\ldots ,n  \label{BHconstr}
\end{equation}%
with generalized forces $Q_{I}=B_{I}^{a}Q_{a}$, in which the $u^{\alpha
},\alpha =1,\ldots \bar{m}$ are set to zero. The terms%
\vspace{-2ex}%
\begin{equation}
\gamma _{IJ}^{a}=\left( \frac{\partial A_{b}^{a}}{\partial q^{c}}-\frac{%
\partial A_{c}^{a}}{\partial q^{b}}\right) B_{I}^{b}B_{J}^{c}
\label{gammaNH}
\end{equation}%
are the Hamel coefficients of the constrained system. The $n-\bar{m}$
equations (\ref{BHconstr}), together with the $n$ kinematic equations%
\begin{equation}
\dot{q}^{a}=B_{I}^{a}u^{I},\ \ I=\bar{m}+1,\ldots ,n  \label{qadot}
\end{equation}%
govern the system dynamics in term of the system state $\left(
q^{a},u^{I}\right) $.

When deriving the Hamel equations (\ref{BHconstr}) for a constrained system,
it is important that the Pfaffian forms $u^{\alpha }$ associated to the
constraints are treated as independent just as the independent
(quasi)velocities $u^{I}$. The $u^{\alpha }$ must be kept during the
derivation in order to obtain the partial derivatives $\frac{\partial 
\mathcal{L}}{\partial u^{\alpha }}$ in (\ref{BHconstr}). In the final
equations, the $u^{\alpha }$ are set to zero (see example in section \ref%
{secBall}). It is also crucial that the original 'unconstrained' Lagrangian
is used in which the $\dot{q}^{a}$ are expressed in terms of all $%
u^{a},a=1,\ldots ,n$ using (\ref{qdu}).

\begin{remark}
\label{remHolonCoord}%
The constraints (\ref{PfaffianConstr}) are holonomic and the velocities $%
u^{I}$ are integrable if and only if 
\begin{equation}
\gamma _{IJ}^{a}\equiv 0.  \label{IntCond3}
\end{equation}%
The fact that the constraints as well as the (quasi)velocities must satisfy
equivalent integrability conditions, encoded in the coefficients (\ref%
{gammaNH}), motivated Boltzmann \cite{Boltzmann1902}, and later Hamel \cite%
{Hamel1904}, to call the $\pi ^{a}$ 'nonholonomic coordinates' (see also 
\cite{Whittaker1904} and \cite{PapastavridisBook2002}) while the
quasi-velocities are called 'nonholonomic' velocities \cite[p. 473]%
{HamelBook},\cite[p. 218]{Sommerfeld1948}. These terms are still used,
although 'nonholonomic coordinates' seems to be an oxymoron. It should be
noted that (\ref{IntCond3}) is the condition for 'complete' nonholonomy.
Even if (\ref{IntCond3}) is not satisfied there may exist lower-dimensional
integral manifolds \cite{BulloLewisBook}.
\end{remark}

\subsubsection{Hamel-Equations in terms of 'holonomic velocities'}

Consider a system described in terms of 'holonomic velocities' $\dot{q}^{a}$
subjected to $\bar{m}$ constraints. The velocities can be partitioned as $%
\dot{\mathbf{q}}=\left( \dot{s}^{\alpha },\dot{r}^{I}\right) $, where $\dot{s%
}^{\alpha },1,\ldots ,\bar{m}$ are dependent, and $\dot{r}^{I},I=\bar{m}%
+1,\ldots ,n$ are independent velocity coordinates, i.e. $u^{I}:=\dot{r}^{I}$%
. Together with the Pfaffian constraints (\ref{PfaffianConstr}), this can be
written in matrix form as%
\begin{equation}
\mathbf{u}=\left( 
\begin{array}{cc}
\mathbf{A}_{1} & \mathbf{A}_{2} \\ 
\mathbf{0} & \mathbf{I}%
\end{array}%
\right) \left( 
\begin{array}{c}
\dot{\mathbf{s}} \\ 
\dot{\mathbf{r}}%
\end{array}%
\right) .  \label{sys1}
\end{equation}%
The solution of (\ref{sys1}) is given by (\ref{u3}) with 
\begin{equation}
\mathbf{B}=\left( 
\begin{array}{cc}
-\mathbf{A}_{1}^{-1} & -\mathbf{A}_{1}^{-1}\mathbf{A}_{2} \\ 
\mathbf{0} & \mathbf{I}%
\end{array}%
\right) .
\end{equation}%
With the particular form of (\ref{sys1}), the Hamel coefficients are%
\begin{eqnarray}
\gamma _{IJ}^{\alpha } &=&\frac{\partial A_{I}^{\alpha }}{\partial r^{J}}-%
\frac{\partial A_{J}^{\alpha }}{\partial r^{I}}+\left( \frac{\partial
A_{\beta }^{\alpha }}{\partial s^{\gamma }}-\frac{\partial A_{\gamma
}^{\alpha }}{\partial s^{\beta }}\right) B_{I}^{\beta }B_{J}^{\gamma }
\label{gammaHolo} \\
&& 
\red%
+\left( \frac{\partial A_{I}^{\alpha }}{\partial s^{\beta }}-\frac{\partial
A_{\beta }^{\alpha }}{\partial r^{I}}\right) B_{J}^{\beta }+\left( \frac{%
\partial A_{\beta }^{\alpha }}{\partial r^{J}}-\frac{\partial A_{J}^{\alpha }%
}{\partial s^{\beta }}\right) B_{I}^{\beta },\ \ \alpha =1,\ldots ,\bar{m}%
\color{black}
\notag
\end{eqnarray}%
and $\gamma _{IJ}^{K}\equiv 0,K=\bar{m}+1,\ldots ,n$.

\begin{remark}
It is instructive noticing that, when using velocities $\dot{q}^{a}$, the
condition (\ref{IntCond3}) are the original Frobenius condition (see \cite%
{Frobenius1875} \S 13-\S 16) for integrability of the constraints (\ref%
{PfaffianConstr})%
\begin{equation}
\left( \frac{\partial A_{\alpha a}}{\partial q^{b}}-\frac{\partial A_{\alpha
b}}{\partial q^{a}}\right) B_{I}^{a}B_{J}^{b}\equiv 0,\ \alpha =1,\ldots ,%
\bar{m}  \label{FrobeniusCond}
\end{equation}%
and the non-vanishing coefficients in (\ref{FrobeniusCond}) are exactly the
Hamel coefficients (\ref{gammaHolo}). The fact that $\gamma _{IJ}^{K}\equiv
0 $ indicates that the $\dot{q}^{a}$ are 'holonomic' velocities. If (\ref%
{FrobeniusCond}) holds, then (\ref{PfaffianConstr}) corresponds to a system
of $m=\bar{m}$ constraints on $q$, whereas for nonholonomic constraints $m<%
\bar{m}$. This is why $\gamma _{IJ}^{\alpha }$ was referred to as the
'object of nonholonomy' \cite{Maisser1997} or 'nonholonomicity' \cite%
{PapastavridisBook1999,PapastavridisBook2002}.
\end{remark}

An alternative way to express the Hamel coefficients, and hence the Hamel
equations, is to first solve the constraints $A_{a}^{\alpha }\dot{q}^{a}=0$
in (\ref{PfaffianConstr}). A solution in terms of the independent $%
\red%
\dot{r}^{I}%
\color{black}%
$ is $\dot{s}^{\alpha }=-\bar{F}_{I}^{\alpha }%
\red%
\dot{r}^{I}%
\color{black}%
,I=\bar{m}+1,\ldots ,n$, where $\bar{F}_{I}^{a}$ is defined by%
\begin{equation}
\dot{\mathbf{q}}=\mathbf{F}\dot{\mathbf{r}},\ ~\mathrm{with~}\mathbf{F}%
:=\left( 
\begin{array}{c}
-\bar{\mathbf{F}} \\ 
\mathbf{I}%
\end{array}%
\right) ,\ \bar{\mathbf{F}}:=\mathbf{A}_{1}^{-1}\mathbf{A}_{2}.  \label{F}
\end{equation}%
Therewith the constraints can be rewritten as%
\begin{equation}
u^{\alpha }=0\text{ with }u^{\alpha }:=\dot{s}^{\alpha }+\bar{F}_{I}^{\alpha
}\dot{r}^{I},\ \ \alpha =1,\ldots ,\bar{m}  \label{uholo}
\end{equation}%
and the corresponding Pfaffian system and its solution are in matrix form%
\begin{equation}
\mathbf{u}=\left( 
\begin{array}{cc}
\mathbf{I} & \bar{\mathbf{F}} \\ 
\mathbf{0} & \mathbf{I}%
\end{array}%
\right) \dot{\mathbf{q}}=\bar{\mathbf{A}}\left( \mathbf{q}\right) \dot{%
\mathbf{q}},\ \ \ \ \dot{\mathbf{q}}=\left( 
\begin{array}{cc}
\mathbf{I} & -\bar{\mathbf{F}} \\ 
\mathbf{0} & \mathbf{I}%
\end{array}%
\right) \mathbf{u}=\bar{\mathbf{B}}\left( \mathbf{q}\right) \mathbf{u}.
\end{equation}%
The Hamel coefficients for this Pfaffian system are $\gamma _{IJ}^{K}\equiv
0,K=\bar{m}+1,\ldots ,n$, and 
\begin{eqnarray}
\gamma _{IJ}^{\alpha } &=&\frac{\partial \bar{A}_{I}^{\alpha }}{\partial
r^{J}}-\frac{\partial \bar{A}_{J}^{\alpha }}{\partial r^{I}}+%
\red%
\frac{\partial \bar{A}_{I}^{\alpha }}{\partial s^{\gamma }}\bar{B}%
_{I}^{\beta }-\frac{\partial \bar{A}_{J}^{\alpha }}{\partial s^{\beta }}\bar{%
B}_{J}^{\gamma }%
\color{black}
\notag \\
&=&\frac{\partial \bar{F}_{I}^{\alpha }}{\partial r^{J}}-\frac{\partial \bar{%
F}_{J}^{\alpha }}{\partial r^{I}}+\frac{\partial \bar{F}_{J}^{\alpha }}{%
\partial s^{\gamma }}\bar{F}_{I}^{\gamma }-\frac{\partial \bar{F}%
_{I}^{\alpha }}{\partial s^{\gamma }}\bar{F}_{J}^{\gamma },\ \ \alpha
=1,\ldots ,\bar{m}.  \label{gammaHolo2}
\end{eqnarray}%
Denote with $\mathcal{L}\left( s^{\alpha },r^{I},u^{a}\right) :=L\left(
q^{a},u^{I}\right) $ the (unconstrained) Lagrangian defined by setting $\dot{%
r}^{I}:=u^{I}$ and $\dot{s}^{\alpha }=u^{\alpha }-\bar{F}_{I}^{\alpha }%
\red%
u^{I}$%
\color{black}%
(the solution of (\ref{uholo})). Noting that $\frac{\partial }{\partial
u^{\alpha }}=\frac{\partial }{\partial \dot{s}^{\alpha }}$, the Hamel
equations (\ref{BHEqu}) of the constrained system 
\red%
in Chaplygin form, for instance, are%
\color{black}%
\begin{equation}
\frac{d}{dt}\frac{\partial \mathcal{L}}{\partial \dot{r}^{I}}-\frac{\partial 
\mathcal{L}}{\partial r^{I}}+\frac{\partial 
\red%
L%
\color{black}%
}{\partial \dot{s}^{\alpha }}\gamma _{IJ}^{\alpha }\dot{r}^{J}=0,\ \ I=\bar{m%
}+1,\ldots ,n.  \label{BHHolo}
\end{equation}%
Now define the constrained 
\red%
Lagrangian 
\color{black}%
$L_{\mathrm{c}}\left( q^{a},\dot{r}^{I}\right) :=L\left( s^{\alpha },r^{I},-%
\bar{F}_{I}^{\alpha }\dot{r}^{I},\dot{r}^{I}\right) $. Since $\frac{\partial 
\mathcal{L}}{\partial \dot{r}^{I}}=\frac{\partial L_{\mathrm{c}}}{\partial 
\dot{r}^{I}}$ the Hamel equations attain the form%
\begin{equation}
\frac{d}{dt}\frac{\partial L_{\mathrm{c}}}{\partial \dot{r}^{I}}-\frac{%
\partial L_{\mathrm{c}}}{\partial r^{I}}+\frac{\partial L_{\mathrm{c}}}{%
\partial s^{\alpha }}\bar{F}_{I}^{\alpha }=-\frac{\partial 
\red%
L%
\color{black}%
}{\partial \dot{s}^{\alpha }}\gamma _{IJ}^{\alpha }\dot{r}^{J},\ \ I=\bar{m}%
+1,\ldots ,n  \label{BHconstrHolo}
\end{equation}%
which were reported in \cite{BlochMarsdenZenkov2005}, where $\bar{F}%
_{I}^{\alpha }$ are interpreted as (kinematic) Ehresmann connection on $T{%
\mathbb{V}}^{n}$ and $\gamma _{IJ}^{\alpha }$ in (\ref{gammaHolo2}) as its
curvature. A similar form of these equations was presented by Chaplygin \cite%
{Chaplygin1897a,PapastavridisBook2002}. It should finally be remarked that
the equations (\ref{BHconstrHolo}) are the explicit form of the equations of
Voronets \cite{Voronets1901} (also spelled Woronetz \cite{Woronetz1910})%
\begin{equation}
\frac{d}{dt}\frac{\partial L}{\partial \dot{r}^{I}}-\frac{\partial L}{%
\partial r^{I}}+F_{I}^{a}\left( \frac{d}{dt}\frac{\partial L}{\partial \dot{s%
}^{\alpha }}-\frac{\partial L}{\partial s^{\alpha }}\right) =F_{I}^{a}Q_{a}
\end{equation}%
for a system subjected to (holonomic or nonholonomic) constraints (\ref%
{PfaffianConstr}).

\begin{remark}
Denote with $F_{J}^{a}$ the components of the orthogonal complement of $%
\mathbf{A}$ in (\ref{F}), so that $\dot{q}^{a}=F_{I}^{a}\dot{r}^{I}$. The
vector fields $F_{I}:=F_{I}^{a}\frac{\partial }{\partial q^{a}}$ define the
constraint distribution $D:=\mathrm{span}~(F_{\bar{m}+1},\ldots ,F_{n})$.
The $\gamma _{IJ}^{a}$ in (\ref{gammaHolo2}) are the coefficients of the Lie
bracket $\left[ F_{I},F_{J}\right] $ in local coordinates. Involutivity of $%
D $ is equivalent to the vanishing of the Hamel coefficients (\ref%
{gammaHolo2}), which is necessary and sufficient for the constraints being
holonomic.
\end{remark}

\begin{remark}
Introduce the one-form $\omega ^{\alpha }:=ds^{\alpha }+\bar{F}_{I}^{\alpha
}dr^{I},\alpha =1,\ldots ,\bar{m}$ so that constraints (\ref{uholo}) are
equivalent to the Pfaffian equations $\omega ^{\alpha }=0$. The exterior
derivative of $\omega ^{\alpha }$ is%
\begin{equation}
d\omega ^{\alpha }=\left( \frac{\partial \bar{F}_{I}^{\alpha }}{\partial
r^{J}}-\frac{\partial \bar{F}_{I}^{\alpha }}{\partial s^{\beta }}\bar{F}%
_{J}^{\beta }\right) dr^{J}\wedge dr^{i}.
\end{equation}%
Contracting $d\omega ^{\alpha }$ with $\dot{\mathbf{q}}=\left( -\bar{F}%
_{I}^{\alpha }\dot{r}^{I},\dot{r}^{I}\right) $ yields $d\omega ^{\alpha
}\left( \dot{\mathbf{q}},\cdot \right) =\gamma _{IJ}^{\alpha }\dot{r}%
^{J}dr^{I}$, and thus 
\begin{equation}
\frac{\partial \mathcal{L}}{\partial \dot{s}^{\alpha }}\gamma _{IJ}^{\alpha }%
\dot{r}^{J}=d\omega ^{\alpha }\left( \dot{\mathbf{q}},\frac{\partial }{%
\partial r^{I}}\right) .
\end{equation}%
This shows again that the $\gamma _{IJ}^{\alpha }$ vanish if and only if the
constraints are holonomic, i.e. $d\omega ^{\alpha }\equiv 0$.
\end{remark}

\subsection{BH-Equations as Euler-Lagrange and Euler-Poincar\'{e} Equations
on $SE\left( 3\right) $}

Hamel \cite{Hamel1904,Hamel-MathAnal1904} already discussed the geometric
properties of the infinitesimal transformations (\ref{dpi}) and when they
form a Lie group, in which case the Hamel coefficients are identical to the
structure constants of this group. In case of the rigid body, this Lie group
is the special Euclidean group $SE\left( 3\right) =SO\left( 3\right) \ltimes 
{\mathbb{R}}^{3}$. The pose of a rigid body is represented by $C\in SE\left(
3\right) $, and its twist by $\mathbf{V}\in se\left( 3\right) $, see
appendix \ref{secSE3}. Important is the invariance of the rigid body
Lagrangian under group actions, i.e. under frame transformations. The
Lagrangian is left-invariant if it is independent of the choice of
body-fixed frame, and it is right-invariant if it is independent from the
choice of inertial frame.

Denote with $l\left( C,\mathbf{V}\right) $ the rigid body Lagrangian in
terms of its pose and twist, which can be considered as a trivialized
version of a Lagrangian $L\left( C,\dot{C}\right) $ defined on $%
T_{C}SE\left( 3\right) $ \cite{MarsdenBook,MarsdenScheurle1993}. The
BH-equations (\ref{BHEqu}) are then%
\begin{equation}
\frac{d}{dt}\frac{\partial l}{\partial V_{i}}\pm \frac{\partial l}{\partial
V_{j}}%
\red%
c_{jik}%
\color{black}%
V_{k}-\frac{\partial l}{\partial \pi _{i}}=W_{i}  \label{EL}
\end{equation}%
with index range $i,j,k=1,\ldots ,6$, where $\mathbf{V},%
\bm{\pi}%
\in se\left( 3\right) ,\mathbf{W}\in se^{\ast }\left( 3\right) $ and the
structure constants are given in (\ref{cijk}). These are the Euler-Lagrange
equations for a left-invariant (+ sign), respectively right-invariant (-
sign), Lagrangian, where $\frac{\partial L}{\partial \pi _{a}}$ is the
expression for $T_{I}L_{C}D_{C}l$, respectively $T_{I}R_{C}D_{C}l$, in local
coordinates, and $D_{C}l\in T^{\ast }G$ denotes the partial derivative of
the Lagrangian $l$ w.r.t. $C\in SE\left( 3\right) $. Note that, since the
basis on $se\left( 3\right) $ is deduced from an orthogonal Cartesian basis,
in the following, it will not be distinguished between co- and contravariant
coordinates, and only subscripts will be used, e.g. $V_{i}=V^{i}$.

If $l$ does not dependent on the pose $C\in SE\left( 3\right) $, then (\ref%
{EL}) become the Euler-Poincar\'{e} (respectively Lie-Poisson) equations%
\begin{equation}
\frac{d}{dt}\frac{\partial l}{\partial V_{i}}\pm 
\red%
c_{jik}%
\color{black}%
\frac{\partial l}{\partial V_{j}}V_{k}=W_{i},\ \ \ \frac{d}{dt}\Pi _{i}\pm 
\red%
c_{jik}%
\color{black}%
\Pi _{j}V_{k}=W_{i}.  \label{Poincare1}
\end{equation}%
with the rigid body momentum $%
\bm{\Pi}%
\in se^{\ast }\left( 3\right) $. They are written in vector form as (note
the change of sign)%
\begin{equation}
\frac{d}{dt}\frac{\partial l}{\partial \mathbf{V}}\mp \mathbf{ad}_{\mathbf{V}%
}^{T}\frac{\partial l}{\partial \mathbf{V}}=\mathbf{W},\ \ \ \frac{d}{dt}%
\bm{\Pi}%
\mp \mathbf{ad}_{\mathbf{V}}^{T}%
\bm{\Pi}%
=\mathbf{W}  \label{Poincare3}
\end{equation}%
where $\mathbf{ad}_{\mathbf{u}}$ is the matrix (\ref{adSE3}) such that $%
\mathbf{ad}_{\mathbf{u}}\mathbf{v}=[\mathbf{u},\mathbf{v}]$. The actual
motion is recovered solving the kinematic reconstruction equations (\ref%
{BodyFixedTwist}) and (\ref{SpatialTwist}), respectively.

More on the equivalence of the original BH-equations and the Euler-Poincar%
\'{e} equations can be found in \cite%
{NeimarkFufaev1972,PapastavridisBook2002}, and in \cite%
{BlochMarsdenZenkov2009}. It was shown in \cite{Guen1968}, for example, that
the variational equations of Volterra and the Poincar\'{e} equations (in
Chetaev form) are equivalent to Hamel's variational equations, where the
connection is made via the transpositional equations (\ref{transpos}).

In the following sections, the Hamel coefficients for the rigid body are
derived using local coordinates as well in a coordinate-free way noting the
geometric meaning of matrix $\mathbf{A}$ in (\ref{u2}), respectively $%
\mathbf{B}$ in (\ref{u3}), which maps to the Lie algebra of $SE\left(
3\right) $.

\section{Hamel Coefficients for the Rotating Rigid Body%
\label{secHamelCoeff}%
}

The rotation of a rigid body is described by the rotation of a body-fixed
frame $\mathcal{F}_{\mathsf{b}}$ relative to a space-fixed inertial frame
(IFR) $\mathcal{F}_{0}$, which is represented by an orthogonal matrix $%
\mathbf{R}=\left( R_{ij}\right) \in SO\left( 3\right) $. This rotation
matrix transforms coordinates of a vector resolved in $\mathcal{F}_{\mathsf{b%
}}$ to its coordinates when resolved in $\mathcal{F}_{0}$, i.e. it
transforms body-fixed to inertial coordinates.

The Hamel coefficients are indeed specific to a particular representation of
the angular velocity. In the following, the angular velocity of the
body-fixed frame $\mathcal{F}_{\mathsf{b}}$ relative to the IFR $\mathcal{F}%
_{0}$ represented in the body-fixed frame (body-fixed representation $%
\bm{\omega}%
^{\mathsf{b}}$) and in the IFR (spatial representation $%
\bm{\omega}%
^{\mathsf{s}}$) are considered, as defined in (\ref{w}) and (\ref{ws1}),
respectively. For both representations, the Hamel coefficients will be
derived. The explicit definition (\ref{gamma}) and the transitory relations
are used to illustrate the application of either method. Rotation matrices
are canonically parametrized in terms of three independent variables
(rotation angles, axis-angle) serving as generalized coordinates. Indices $%
i,j,k,l=1,2,3$ refer to to the Cartesian basis, indices $a,b,c=1,2,3$ refer
to the generalized coordinates used to describe rotations. Since vectors are
resolved in orthogonal Cartesian bases, it will not be distinguished between
co- and contravariant coordinates, and the corresponding indices are written
as subscripts.

\subsection{Derivation from the explicit definition%
\label{secGammaRot}%
}

The angular velocity serves as quasi-velocity. The Hamel coefficients are
now derived from the explicit definition (\ref{gamma}) for the body-fixed
representation of angular velocity. For spatial rotations, the matrices $%
\mathbf{A}$ and $\mathbf{B}$ are placed by $%
\bm{\Omega}%
$ in (\ref{w}) and $%
\bm{\Psi}%
$ in (\ref{Psi}), respectively (superscript \textsf{b} will be omitted).
Thus (\ref{gamma}) becomes%
\begin{equation}
\gamma _{ij}^{k}=\left( \frac{\partial \Omega _{ka}}{\partial \varphi ^{b}}-%
\frac{\partial \Omega _{kb}}{\partial \varphi ^{a}}\right) \Psi _{i}^{a}\Psi
_{j}^{b},\ \ i,j,k=1,2,3.  \label{IntCond2}
\end{equation}%
For fixed $k$, a skew symmetric tensor can be associated to $\gamma
_{ij}^{k} $, which is expressed with (\ref{Aij}) as%
\begin{equation}
\varepsilon _{rks}\gamma _{ij}^{k}=\varepsilon _{rks}\left( \frac{\partial
\Omega _{ka}}{\partial \varphi ^{b}}-\frac{\partial \Omega _{kb}}{\partial
\varphi ^{a}}\right) \Psi _{i}^{a}\Psi _{j}^{b}=\left( \frac{\partial
^{2}\omega _{rs}}{\partial \varphi ^{b}\partial \dot{\varphi}^{a}}-\frac{%
\partial ^{2}\omega _{rs}}{\partial \dot{\varphi}^{a}\partial \varphi ^{b}}%
\right) \Psi _{i}^{a}\Psi _{j}^{b}.
\end{equation}%
Using (\ref{wij1}), the partial derivatives of the first term in the bracket
of the last term is%
\begin{equation}
\frac{\partial ^{2}\omega _{rs}}{\partial \varphi ^{b}\partial \dot{\varphi}%
^{a}}=R_{lr}\frac{\partial ^{2}R_{ls}}{\partial \varphi ^{a}\partial \varphi
^{b}}+\frac{\partial R_{lr}}{\partial \varphi ^{b}}\frac{\partial R_{ls}}{%
\partial \varphi ^{a}}  \label{d2wij}
\end{equation}%
and analogously for the second term by swapping the indices $b$ and $a$, and
thus%
\begin{equation}
\varepsilon _{rks}\gamma _{ij}^{k}=\left( \frac{\partial R_{lr}}{\partial
\varphi ^{b}}\frac{\partial R_{ls}}{\partial \varphi ^{a}}-\frac{\partial
R_{lr}}{\partial \varphi ^{a}}\frac{\partial R_{ls}}{\partial \varphi ^{b}}%
\right) \Psi _{i}^{a}\Psi _{j}^{b}.
\end{equation}%
Replacing the partial derivatives with the relation (\ref{wIdent}) yields%
\begin{equation}
\varepsilon _{rks}\gamma _{ij}^{k}=R_{lk}\varepsilon _{kjr}R_{lp}\varepsilon
_{pis}-R_{lp}\varepsilon _{pir}R_{lk}\varepsilon _{kjs}=\varepsilon
_{kjr}\varepsilon _{kis}-\varepsilon _{kir}\varepsilon _{kjs}=\varepsilon
_{rks}\varepsilon _{kij}
\end{equation}%
where the last term follows with $\varepsilon _{kis}\varepsilon
_{kjr}-\varepsilon _{kir}\varepsilon _{kjs}=\delta _{ij}\delta _{rs}-\delta
_{ir}\delta _{js}-\delta _{ij}\delta _{rs}+\delta _{is}\delta _{jr}=\delta
_{is}\delta _{jr}-\delta _{ir}\delta _{js}=\varepsilon _{ksr}\varepsilon
_{kij}=\varepsilon _{rps}\varepsilon _{pij}$. The Hamel coefficients for the
rotating rigid body in terms of body-fixed angular velocity are thus%
\begin{equation}
\gamma _{ij}^{k}=\varepsilon _{kij}.  \label{gammaOmegaBody}
\end{equation}%
An analog derivation for the spatial representation yields the Hamel
coefficients $\gamma _{ij}^{k}=-\varepsilon _{kij}$.

\subsection{Derivation via the transitory relation%
\label{secTransRot}%
}

The Hamel coefficients can be found directly from the transitory relation (%
\ref{transpos}), instead of explicitly evaluating the definition (\ref{gamma}%
) in terms of local coordinates on $SO\left( 3\right) $. It will turn out
that this is a straightforward and simple approach. Hamel already pointed
out that it might be easier to derive the Euler-Lagrange equations by
directly evaluating the transitory relations. He used vectorial relations
for derivatives of the angular velocity, and avoided explicit coordinate
expressions (see \cite[p. 488]{HamelBook} and also the discussion in \cite[%
p. 374]{PapastavridisBook2002}).

\subsubsection{Angular velocity in body-fixed representation}

Denote with $%
\bm{\theta}%
^{\mathsf{b}}\in {\mathbb{R}}^{3}$ the vector of quasi-coordinates
associated to the rotational motion according to $d%
\bm{\theta}%
^{\mathsf{b}}=%
\bm{\omega}%
^{\mathsf{b}}dt$, which can be regarded as 'quasi-rotation vector'
represented in the moving body-fixed frame (but it is not a vector along the
instantaneous rotation axis). The differential of these quasi-coordinates
can be expressed using the definition (\ref{w1}) of the body-fixed angular
velocity as%
\begin{equation}
d\widetilde{%
\bm{\theta}%
}^{\mathsf{b}}=\widetilde{%
\bm{\omega}%
}^{\mathsf{b}}dt=\mathbf{R}^{T}\frac{\partial \mathbf{R}}{\partial \varphi
^{a}}d\varphi ^{a}=\mathbf{R}^{T}d\mathbf{R}  \label{dR}
\end{equation}%
and analogously the variation as $\delta \widetilde{%
\bm{\theta}%
}^{\mathsf{b}}=\mathbf{R}^{T}\delta \mathbf{R}$. The subscript $\mathsf{b}$
will be omitted in the following. In case of spatial rotations, the
transitory relation (\ref{transpos}) is%
\begin{equation}
\frac{d}{dt}\left( d\delta \theta _{k}-\delta d\theta _{k}\right) =\gamma
_{ij}^{k}\omega _{j}\delta \theta _{i}.  \label{transposeSO3b}
\end{equation}%
The bracketed term on the left reads in vector form $d\delta 
\bm{\theta}%
-\delta d%
\bm{\theta}%
$. Writing the latter as skew symmetric matrix, and using (\ref{dR}) yields%
\begin{eqnarray}
d\delta \widetilde{%
\bm{\theta}%
}-\delta d\widetilde{%
\bm{\theta}%
} &=&d\left( \mathbf{R}^{T}\delta \mathbf{R}\right) -\delta \left( \mathbf{R}%
^{T}d\mathbf{R}\right)  \notag \\
&=&d\mathbf{R}^{T}\delta \mathbf{R}+\mathbf{R}^{T}d\delta \mathbf{R}-\delta 
\mathbf{R}^{T}d\mathbf{R-R}^{T}\delta d\mathbf{R}  \notag \\
&=&-\mathbf{R}^{T}d\mathbf{RR}^{T}\delta \mathbf{R}+\mathbf{R}^{T}d\delta 
\mathbf{R}+\mathbf{R}^{T}\delta \mathbf{RR}^{T}d\mathbf{R-R}^{T}\delta d%
\mathbf{R}  \notag \\
&=&\delta \widetilde{%
\bm{\theta}%
}d\widetilde{%
\bm{\theta}%
}-d\widetilde{%
\bm{\theta}%
}\delta \widetilde{%
\bm{\theta}%
}+\mathbf{R}^{T}\left( d\delta \mathbf{R}-\delta d\mathbf{R}\right) .
\end{eqnarray}%
Noting that the term $d\delta \mathbf{R}-\delta d\mathbf{R}$ vanishes
identically, this simplifies to%
\begin{equation}
d\delta \widetilde{%
\bm{\theta}%
}-\delta d\widetilde{%
\bm{\theta}%
}=\delta \widetilde{%
\bm{\theta}%
}d\widetilde{%
\bm{\theta}%
}-d\widetilde{%
\bm{\theta}%
}\delta \widetilde{%
\bm{\theta}%
}=[\delta \widetilde{%
\bm{\theta}%
},d\widetilde{%
\bm{\theta}%
}].  \label{bodyfixed1}
\end{equation}%
Transforming (\ref{bodyfixed1}) to vector form, using (\ref{skew}), and
using the vector form of the Lie bracket (\ref{SO3Lie}), yields $\left(
d\delta 
\bm{\theta}%
-\delta d%
\bm{\theta}%
\right) =\mathbf{ad}_{\delta 
\bm{\theta}%
}d%
\bm{\theta}%
=\delta \widetilde{%
\bm{\theta}%
}d%
\bm{\theta}%
$. The vector form of the transitory relation (\ref{transposeSO3b}) follows
with $%
\bm{\omega}%
=\frac{d%
\bm{\theta}%
}{dt}$ as $\frac{d}{dt}\left( d\delta 
\bm{\theta}%
-\delta d%
\bm{\theta}%
\right) =\mathbf{ad}_{\delta 
\bm{\theta}%
}%
\bm{\omega}%
$. With (\ref{SO3Lie2}), the components of this relation are%
\begin{equation}
\frac{d}{dt}\left( d\delta \theta _{k}-\delta d\theta _{k}\right)
=\varepsilon _{kij}\delta \theta _{i}\omega _{j}.  \label{bodyfixed3}
\end{equation}%
Comparing (\ref{bodyfixed3}) and (\ref{transposeSO3b}) confirms that $\gamma
_{ij}^{k}=\varepsilon _{kij}$, in accordance to (\ref{gammaOmegaBody}).
Clearly, the derivation of the Hamel coefficients via the transitory
relations is simpler than using the explicit definition (\ref{gamma}).

\subsubsection{Angular velocity in spatial representation%
\label{secSpatialOmega}%
}

The vector $%
\bm{\theta}%
^{\mathsf{s}}$ of quasi-coordinates ('quasi-rotation vector'), represented
in IFR, is defined by $d%
\bm{\theta}%
^{\mathsf{s}}=%
\bm{\omega}%
^{\mathsf{s}}dt$ with spatial representation of the angular velocity (\ref%
{ws1}). It thus satisfies the relations $d\widetilde{%
\bm{\theta}%
}^{\mathsf{s}}=d\mathbf{RR}^{T}$ and $\delta \widetilde{%
\bm{\theta}%
}^{\mathsf{s}}=\delta \mathbf{R\mathbf{R}}^{T}$. The transitory relation (%
\ref{transposeSO3b}) applies in terms of $%
\bm{\theta}%
^{\mathsf{s}}$, and the derivation of the Hamel coefficients proceeds as
above. The subscript \textsf{s} will again be omitted. Noting again that $%
d\delta \mathbf{R}-\delta d\mathbf{R}\equiv \mathbf{0}$, yields%
\begin{eqnarray}
d\delta \widetilde{%
\bm{\theta}%
}-\delta d\widetilde{%
\bm{\theta}%
} &=&d\left( \delta \mathbf{RR}^{T}\right) -\delta \left( d\mathbf{RR}%
^{T}\right)  \notag \\
&=&d\delta \mathbf{RR}^{T}-\delta \mathbf{RR}^{T}d\mathbf{RR}^{T}-\delta d%
\mathbf{RR}^{T}+d\mathbf{RR}^{T}\delta \mathbf{RR}^{T}  \notag \\
&=&\left( d\delta \mathbf{R}-\delta d\mathbf{R}\right) \mathbf{R}^{T}+d%
\widetilde{%
\bm{\theta}%
}\delta \widetilde{%
\bm{\theta}%
}-\delta \widetilde{%
\bm{\theta}%
}d\widetilde{%
\bm{\theta}%
}  \notag \\
&=&[d\widetilde{%
\bm{\theta}%
},\delta \widetilde{%
\bm{\theta}%
}],  \label{transRot}
\end{eqnarray}%
which is equivalent to the vector form $\frac{d}{dt}\left( d\delta 
\bm{\theta}%
-\delta d%
\bm{\theta}%
\right) =-\mathbf{ad}_{\delta 
\bm{\pi}%
}%
\bm{\omega}%
$. The coordinate form (\ref{SO3Lie2}) of the Lie bracket yields%
\begin{equation}
\frac{d}{dt}\left( d\delta \theta _{k}-\delta d\theta _{k}\right)
=-\varepsilon _{kij}\delta \theta _{i}\omega _{j}.  \label{spatialtmp}
\end{equation}%
Comparing (\ref{spatialtmp}) with (\ref{transposeSO3b}) in terms of the
spatial representation, shows that the Hamel coefficients are 
\begin{equation}
\gamma _{ij}^{k}=-\varepsilon _{kij},  \label{gammaWs}
\end{equation}%
which have the opposite sign as those when using the angular velocity in
body-fixed representation.

\section{Boltzmann-Hamel Equations for Rigid Body with Fixed Point%
\label{secBHERot}%
}

\subsection{Body-fixed angular velocity}

For simplicity, no potential forces are included, so that the Lagrangian is
identical to the kinetic energy. Denote with $%
\bm{\Theta}%
^{\mathsf{b}}$ the constant inertia tensor w.r.t. the center of rotation
represented in the body-fixed frame $\mathcal{F}_{\mathsf{b}}$. The kinetic
energy of the rotating body in terms of the angular velocity in body-fixed
representation is%
\begin{equation}
T(%
\bm{\omega}%
^{\mathsf{b}})=\frac{1}{2}\omega _{i}^{\mathsf{b}}\Theta _{ij}^{\mathsf{b}%
}\omega _{j}^{\mathsf{b}}=\frac{1}{2}%
\bm{\omega}%
^{\mathsf{b}T}%
\bm{\Theta}%
^{\mathsf{b}}%
\bm{\omega}%
^{\mathsf{b}}.  \label{TRotb}
\end{equation}

Since $T$ does not depend on the orientation, the BH-equations (\ref{BHEqu})
simplify to%
\begin{equation}
\frac{d}{dt}\frac{\partial T}{\partial \omega _{i}^{\mathsf{b}}}+\frac{%
\partial T}{\partial \omega _{j}^{\mathsf{b}}}\gamma _{ik}^{j}\omega _{k}^{%
\mathsf{b}}=\tau _{i}^{\mathsf{b}}
\end{equation}%
where $\tau _{i}^{\mathsf{b}}$ are components of the applied torque resolved
in $\mathcal{F}_{\mathsf{b}}$. With $\frac{\partial T}{\partial \omega _{i}^{%
\mathsf{b}}}=\Theta _{ij}^{\mathsf{b}}\omega _{j}^{\mathsf{b}}$ and with the
Hamel coefficients $\gamma _{ik}^{j}=\varepsilon _{jik}=\varepsilon _{ikj}$
in (\ref{gammaOmegaBody}), the BH-equations are%
\begin{equation}
\Theta _{ij}^{\mathsf{b}}\dot{\omega}_{j}^{\mathsf{b}}+\varepsilon
_{ikj}\omega _{k}^{\mathsf{b}}\Theta _{jl}^{\mathsf{b}}\omega _{l}^{\mathsf{b%
}}=\tau _{i}^{\mathsf{b}}.  \label{EulerBodyFixed}
\end{equation}%
These are indeed the dynamic Euler-equations for a rotating rigid body with
fixed point in terms of body-fixed angular velocity, which are the
Euler-Poincar\'{e} equations (\ref{Poincare3}) on $SO\left( 3\right) $, with
the structure constants $\gamma _{ij}^{k}=\varepsilon _{kij}$, for the
left-invariant kinetic energy (\ref{TRotb}).. They attain the well-known
matrix form%
\begin{equation}
\bm{\Theta}%
^{\mathsf{b}}\dot{%
\bm{\omega}%
}^{\mathsf{b}}+\widetilde{%
\bm{\omega}%
}{^{\mathsf{b}}}%
\bm{\Theta}%
^{\mathsf{b}}%
\bm{\omega}%
^{\mathsf{b}}=\mathbf{\tau }^{\mathsf{b}}.
\end{equation}

\subsection{Spatial angular velocity}

The kinetic energy of the rotating body is%
\begin{equation}
T(%
\bm{\varphi}%
,%
\bm{\omega}%
^{\mathsf{s}})=\frac{1}{2}\omega _{i}^{\mathsf{s}}\Theta _{ij}^{\mathsf{s}%
}\omega _{j}^{\mathsf{s}}=\frac{1}{2}%
\bm{\omega}%
^{\mathsf{s}T}%
\bm{\Theta}%
^{\mathsf{s}}%
\bm{\omega}%
^{\mathsf{s}}  \label{TRots}
\end{equation}%
where $%
\bm{\Theta}%
^{\mathsf{s}}\left( 
\bm{\varphi}%
\right) $ is the inertia tensor w.r.t. to the fixed point expressed in the
IFR, and $\varphi ^{a}$ are local coordinates on $SO\left( 3\right) $. The
spatial inertia tensor depends on the orientation and is related to the
(constant) body-fixed inertia tensor as $%
\bm{\Theta}%
^{\mathsf{s}}=\mathbf{R}%
\bm{\Theta}%
^{\mathsf{b}}\mathbf{R}^{T}$.

The BH-equations (\ref{BHEqu2}) in terms of spatial angular velocity, and
with the Hamel coefficients $\gamma _{ik}^{j}=-\varepsilon _{jik}$ in (\ref%
{gammaWs}), are%
\begin{equation}
\frac{d}{dt}\frac{\partial T}{\partial \omega _{i}^{\mathsf{s}}}-\frac{%
\partial T}{\partial \varphi ^{a}}\Psi _{i}^{a,\mathsf{s}}+\frac{\partial T}{%
\partial \omega _{j}^{\mathsf{s}}}\gamma _{ik}^{j}\omega _{k}^{\mathsf{s}%
}=\tau _{i}^{\mathsf{s}}
\end{equation}%
where $\tau _{i}$ are the components of the applied torque, and $\Psi
_{j}^{a,\mathsf{s}}$ is the inverse of $\Omega _{ia}^{\mathsf{s}}$ in (\ref%
{ws1}).

From $\frac{\partial T}{\partial \omega _{i}^{\mathsf{s}}}=\Theta _{ij}^{%
\mathsf{s}}\omega _{j}^{\mathsf{s}}$, and using the relation $\dot{%
\bm{\Theta}%
}^{\mathsf{s}}=\dot{\mathbf{R}}%
\bm{\Theta}%
^{\mathsf{b}}\mathbf{R}^{T}-\mathbf{R}%
\bm{\Theta}%
^{\mathsf{b}}\mathbf{R}^{T}\dot{\mathbf{R}}\mathbf{R}^{T}=\widetilde{%
\bm{\omega}%
}^{\mathsf{s}}%
\bm{\Theta}%
^{\mathsf{s}}-%
\bm{\Theta}%
^{\mathsf{s}}\widetilde{%
\bm{\omega}%
}^{\mathsf{s}}$, yields 
\begin{equation}
\frac{d}{dt}\frac{\partial T}{\partial \omega _{i}^{\mathsf{s}}}=\Theta
_{ij}^{\mathsf{s}}\dot{\omega}_{j}^{\mathsf{s}}+\dot{\Theta}_{ij}\omega
_{j}^{\mathsf{s}}=\Theta _{ij}^{\mathsf{s}}\dot{\omega}_{j}^{\mathsf{s}%
}+\varepsilon _{ikl}\omega _{k}^{\mathsf{s}}\Theta _{lj}^{\mathsf{s}}\omega
_{j}^{\mathsf{s}}.  \label{BHRotSpatial}
\end{equation}%
The partial derivative of $%
\bm{\Theta}%
^{\mathsf{s}}$ is found with (\ref{ws1}) as%
\begin{eqnarray}
\frac{\partial 
\bm{\Theta}%
^{\mathsf{s}}}{\partial \varphi ^{a}} &=&\frac{\partial \mathbf{R}}{\partial
\varphi ^{a}}%
\bm{\Theta}%
^{\mathsf{b}}\mathbf{R}^{T}-\mathbf{R}%
\bm{\Theta}%
^{\mathsf{b}}\mathbf{R}^{T}\frac{\partial \mathbf{R}}{\partial \varphi ^{a}}%
\mathbf{R}^{T}=\widetilde{%
\bm{\Omega}%
}_{a}^{\mathsf{s}}\mathbf{R}%
\bm{\Theta}%
^{\mathsf{b}}\mathbf{R}^{T}-\mathbf{R}%
\bm{\Theta}%
^{\mathsf{b}}\mathbf{R}^{T}\widetilde{%
\bm{\Omega}%
}_{a}^{\mathsf{s}}  \notag \\
&=&\widetilde{%
\bm{\Omega}%
}_{a}^{\mathsf{s}}%
\bm{\Theta}%
^{\mathsf{s}}-%
\bm{\Theta}%
^{\mathsf{s}}\widetilde{%
\bm{\Omega}%
}_{a}^{\mathsf{s}}.  \label{partial1}
\end{eqnarray}%
This gives rise to the partial derivatives of the kinetic energy (omitting
superscript \textsf{s}) 
\begin{eqnarray}
\frac{\partial T}{\partial \varphi ^{a}} &=&\frac{1}{2}\omega _{j}\frac{%
\partial \Theta _{jk}}{\partial \varphi ^{a}}\omega _{k}=\frac{1}{2}\left(
\omega _{j}\varepsilon _{jls}\Omega _{la}\Theta _{sk}\omega _{k}-\omega
_{j}\Theta _{js}\varepsilon _{slk}\Omega _{la}\omega _{k}\right)  \notag \\
&=&-\Omega _{ja}\varepsilon _{jsl}\omega _{s}\Theta _{lk}\omega _{k}.
\label{partial2}
\end{eqnarray}%
Finally, inserting (\ref{partial1}) and (\ref{partial2}) into (\ref%
{BHRotSpatial}) yields%
\begin{equation}
\Theta _{ij}\dot{\omega}_{j}+\varepsilon _{ikl}\omega _{k}\Theta _{lj}\omega
_{j}+\Psi _{i}^{a}\Omega _{ka}\varepsilon _{ksl}\omega _{s}\Theta
_{lr}\omega _{r}-\varepsilon _{jik}\Theta _{js}\omega _{s}\omega _{k}=\tau
_{i}.  \label{BHRotSpatial2}
\end{equation}%
Noting that $\Psi _{i}^{a}\Omega _{ka}=\delta _{ik}$ and $\varepsilon
_{ikl}\omega _{k}\Theta _{lj}\omega _{j}=\varepsilon _{jik}\Theta
_{js}\omega _{s}\omega _{k}$, the dynamic Euler-equations for the rigid body
rotating about a fixed point in terms of the spatial angular velocity are%
\begin{equation}
\Theta _{ij}^{\mathsf{s}}\dot{\omega}_{j}^{\mathsf{s}}+\varepsilon
_{ikl}\omega _{k}^{\mathsf{s}}\Theta _{lj}^{\mathsf{s}}\omega _{j}^{\mathsf{s%
}}=\tau _{i}^{\mathsf{s}}.  \label{EulerSpatial}
\end{equation}%
These are the Euler-Lagrange equations (\ref{EL}) w.r.t. the right-invariant
kinetic energy (\ref{TRots}).

The equations (\ref{EulerSpatial}) and (\ref{EulerBodyFixed}) are
structurally identical, but clearly their derivation from the kinetic energy
is different. Moreover, (\ref{EulerSpatial}) can be derived from spatial
momentum conservation as%
\begin{equation}
\dot{L}_{j}^{\mathsf{s}}=\tau _{i}^{\mathsf{s}}
\end{equation}%
where $L_{i}:=\Theta _{ij}^{\mathsf{s}}\omega _{j}^{\mathsf{s}}$ is the
spatial angular momentum.

\section{Hamel Coefficients for the Spatially Moving Rigid Body%
\label{secHamelCoeffFree}%
}

\subsection{Derivation from the explicit definition}

For sake of brevity, in the following only the body-fixed representation of
rigid body twists is considered, as defined in (\ref{BodyFixedTwist}). The
derivation for the spatial representation proceeds analogously.

The derivation proceeds as for the rotating rigid body in sec. \ref%
{secGammaRot}. The superscript $\mathsf{b}$ of $V^{\mathsf{b}}$ indicating
body-fixed representation will be omitted. The Hamel coefficients $\gamma
_{ij}^{k},i,j,k=1,\ldots ,6$ are written in form of an $se\left( 3\right) $%
-tensor $h_{\alpha k\beta }\gamma _{ij}^{k},\alpha ,\beta =1,\ldots ,4$.
Inserting the relation (\ref{Va}) for the body-fixed twist in the definition
(\ref{gamma}), and using (\ref{Vab3}) and (\ref{Vab2}) yields%
\begin{eqnarray}
h_{\alpha k\beta }\gamma _{ij}^{k} &=&h_{\alpha i\beta }\left( \frac{%
\partial J_{ka}}{\partial q^{b}}-\frac{\partial J_{kb}}{\partial q^{a}}%
\right) G_{i}^{a}G_{j}^{b} \\
&=&\left( \frac{\partial ^{2}V_{\alpha \beta }}{\partial q^{b}\partial \dot{q%
}^{a}}-\frac{\partial ^{2}V_{\alpha \beta }}{\partial q^{a}\partial \dot{q}%
^{b}}\right) G_{i}^{a}G_{j}^{b}.
\end{eqnarray}%
With (\ref{Vab3}), the mixed partial derivatives of the twist tensor are%
\begin{equation}
\frac{\partial ^{2}V_{\alpha \beta }}{\partial q^{b}\partial \dot{q}^{a}}%
=D_{\alpha \sigma }\frac{\partial ^{2}C_{\sigma \beta }}{\partial
q^{b}\partial q^{a}}+\frac{\partial D_{\alpha \sigma }}{\partial q^{b}}\frac{%
\partial C_{\sigma \beta }}{\partial q^{a}}.  \label{d2V}
\end{equation}%
Noting that the second-order derivatives commute, this leads to%
\begin{equation}
h_{\alpha k\beta }\gamma _{ij}^{k}=\left( \frac{\partial D_{\alpha \sigma }}{%
\partial q^{b}}\frac{\partial C_{\sigma \beta }}{\partial q^{a}}-\frac{%
\partial D_{\alpha \sigma }}{\partial q^{a}}\frac{\partial C_{\sigma \beta }%
}{\partial q^{b}}\right) G_{i}^{a}G_{j}^{b}.  \label{hgamma}
\end{equation}%
Inserting (\ref{wIdentV}) into (\ref{hgamma}), and noting that $D_{\mu
\sigma }C_{\sigma \nu }=\delta _{\mu \nu }$, yields 
\begin{eqnarray}
h_{\alpha i\beta }\gamma _{ij}^{k} &=&h_{\alpha j\mu }D_{\mu \sigma
}C_{\sigma \nu }h_{\nu i\beta }-h_{\alpha i\mu }D_{\mu \sigma }C_{\sigma \nu
}h_{\nu j\beta }  \notag \\
&=&h_{\alpha j\nu }h_{\nu i\beta }-h_{\alpha i\nu }h_{\nu j\beta }  \notag \\
&=&h_{\alpha k\beta }c_{kij}  \label{gammaVb}
\end{eqnarray}%
where the identity (\ref{hc}) is used. The Hamel coefficients $\gamma
_{ij}^{k}=c_{kij}$, for the spatially moving rigid body are thus the
structure constants (\ref{cijk}) of $SE\left( 3\right) $, when the
body-fixed twist is used as quasi-velocity.

\subsection{Derivation via the transitory relation}

\subsubsection{Body-fixed representation of rigid body twists%
\label{secBHCoeffBodyFixed}%
}

Instead of explicitly evaluating the definition (\ref{gamma}), the Hamel
coefficients are derived from the transitory relation (\ref{transpos}), as
in sec. \ref{secTransRot}. Denote with $%
\bm{\pi}%
^{\mathsf{b}}=(%
\bm{\theta}%
^{\mathsf{b}},%
\bm{\rho}%
^{\mathsf{b}})\in {\mathbb{R}}^{6}$ the vector of quasi-coordinates
associated to the rigid body motion so that $d%
\bm{\pi}%
^{\mathsf{b}}=\mathbf{V}^{\mathsf{b}}dt$, which can be regarded as
'quasi-screw coordinates' expressed in the body-fixed frame. With the
definition (\ref{BodyFixedTwist}) of the body-fixed twist, the differential
and variation of these quasi-coordinates are expressed as%
\begin{equation}
\delta \widehat{%
\bm{\pi}%
}^{\mathsf{b}}=\mathbf{C}^{-1}\delta \mathbf{C}=\left( 
\begin{array}{cc}
\mathbf{R}^{T}\delta \mathbf{R} & \mathbf{R}^{T}\delta \mathbf{r} \\ 
\mathbf{0} & 0%
\end{array}%
\right) =\left( 
\begin{array}{cc}
\delta \widetilde{%
\bm{\theta}%
}^{\mathsf{b}} & \delta 
\bm{\rho}%
^{\mathsf{b}} \\ 
\mathbf{0} & 0%
\end{array}%
\right) \mathbf{,}\ \ d\widehat{%
\bm{\pi}%
}^{\mathsf{b}}=\mathbf{C}^{-1}d\mathbf{C}=\left( 
\begin{array}{cc}
\mathbf{R}^{T}d\mathbf{R} & \mathbf{R}^{T}d\mathbf{r} \\ 
\mathbf{0} & 0%
\end{array}%
\right) .  \label{dC}
\end{equation}%
The term $d\widetilde{%
\bm{\theta}%
}^{\mathsf{b}}=\mathbf{R}^{T}d\mathbf{R}$ defines the differential $d%
\bm{\theta}%
^{\mathsf{b}}$ of the 'quasi-rotation vector' in (\ref{dR}), and $d%
\bm{\rho}%
^{\mathsf{b}}=\mathbf{R}^{T}d\mathbf{r}$ is the differential of the position
vector of the body-fixed frame expressed in the body-fixed frame. Again, in
the following, the subscript $\mathsf{b}$ will be omitted.

Written in vector form, the transitory relation (\ref{transpos}) involves
the term $d\delta 
\bm{\pi}%
-\delta d%
\bm{\pi}%
$, which is equivalent to the matrix form $d\delta \widehat{%
\bm{\pi}%
}-\delta d\widehat{%
\bm{\pi}%
}$ using the isomorphism (\ref{hat}). Inserting (\ref{dC}) yields%
\begin{eqnarray}
d\delta \widehat{%
\bm{\pi}%
}-\delta d\widehat{%
\bm{\pi}%
} &=&d\left( \mathbf{C}^{-1}\delta \mathbf{C}\right) -\delta \left( \mathbf{C%
}^{-1}d\mathbf{C}\right)  \notag \\
&=&d\mathbf{C}^{-1}\delta \mathbf{C}+\mathbf{C}^{-1}d\delta \mathbf{C}%
-\delta \mathbf{C}^{-1}d\mathbf{C-C}^{-1}\delta d\mathbf{C}  \notag \\
&=&-\mathbf{C}^{-1}d\mathbf{CC}^{-1}\delta \mathbf{C}+\mathbf{C}^{-1}d\delta 
\mathbf{C}+\mathbf{C}^{-1}\delta \mathbf{CC}^{-1}d\mathbf{C-C}^{-1}\delta d%
\mathbf{C}  \notag \\
&=&\delta \widehat{%
\bm{\pi}%
}d\widehat{%
\bm{\pi}%
}-d\widehat{%
\bm{\pi}%
}\delta \widehat{%
\bm{\pi}%
}+\mathbf{C}^{-1}\left( d\delta \mathbf{C}-\delta d\mathbf{C}\right)  \notag
\\
&=&\delta \widehat{%
\bm{\pi}%
}d\widehat{%
\bm{\pi}%
}-d\widehat{%
\bm{\pi}%
}\delta \widehat{%
\bm{\pi}%
}=[\delta \widehat{%
\bm{\pi}%
},d\widehat{%
\bm{\pi}%
}].  \label{transBody}
\end{eqnarray}%
The Lie bracket, and thus (\ref{transBody}), can be written in vector form
using (\ref{adSE3}). Making use of the skew symmetry of the Lie bracket, (%
\ref{transBody}) is equivalent to $d\delta 
\bm{\pi}%
-\delta d%
\bm{\pi}%
=-\mathbf{ad}_{d%
\bm{\pi}%
}\delta 
\bm{\pi}%
$. The transitory relation (\ref{transpos}) thus becomes, with $d%
\bm{\pi}%
=\mathbf{V}dt$, 
\begin{equation}
\frac{d}{dt}\left( d\delta 
\bm{\pi}%
-\delta d%
\bm{\pi}%
\right) =-\mathbf{ad}_{\mathbf{V}}\delta 
\bm{\pi}%
=-\left( 
\begin{array}{c}
\widetilde{%
\bm{\omega}%
}\delta 
\bm{\theta}
\\ 
\widetilde{%
\bm{\omega}%
}\delta 
\bm{\rho}%
+\widetilde{\mathbf{v}}\delta 
\bm{\theta}%
\end{array}%
\right) .  \label{transBody3}
\end{equation}%
In components, (\ref{transBody3}) reads 
\begin{eqnarray}
\frac{d}{dt}\left( d\delta \pi _{k}-\delta d\pi _{k}\right)
&=&-c_{kij}V_{i}\delta \pi _{j}  \label{transBody24} \\
&=&c_{kij}\delta \pi _{i}V_{j}.  \label{transBody2}
\end{eqnarray}%
Comparing (\ref{transBody2}) and (\ref{transpos}) shows again that the Hamel
coefficients $\gamma _{ij}^{k}=c_{kij}$ are the structure constants of $%
SE\left( 3\right) $. The above derivation is much simpler than using the
definition of Hamel coefficients (\ref{gamma}) as was already mentioned by
Hamel. Moreover, the above matrix representation of $SE\left( 3\right) $
further simplifies the derivation.

\subsubsection{Spatial representation of rigid body twists}

The vector of quasi-coordinates $%
\bm{\pi}%
^{\mathsf{s}}$ ('quasi-screw coordinates' represented in IFR) is introduced
so that $d%
\bm{\pi}%
^{\mathsf{s}}=\mathbf{V}^{\mathsf{s}}dt$. It is determined by the expression
(\ref{SpatialTwist}) for the rigid body twist in spatial representation as%
\begin{equation}
\delta \widehat{%
\bm{\pi}%
}^{\mathsf{s}}=\delta \mathbf{CC}^{-1}=\left( 
\begin{array}{cc}
\delta \mathbf{RR}^{T} & \delta \mathbf{r}-\delta \widetilde{%
\bm{\theta}%
}\mathbf{r} \\ 
\mathbf{0} & 0%
\end{array}%
\right) =\left( 
\begin{array}{cc}
\delta \widetilde{%
\bm{\theta}%
}^{\mathsf{s}} & \delta 
\bm{\rho}%
^{\mathsf{s}} \\ 
\mathbf{0} & 0%
\end{array}%
\right) \mathbf{,}\ \ d\widehat{%
\bm{\pi}%
}^{\mathsf{s}}=d\mathbf{CC}^{-1}=\left( 
\begin{array}{cc}
d\mathbf{RR}^{T} & d\mathbf{rR}^{T} \\ 
\mathbf{0} & 0%
\end{array}%
\right)  \label{dCCT}
\end{equation}%
where $\delta \widetilde{%
\bm{\theta}%
}^{\mathsf{s}}=\delta \mathbf{RR}^{T}$ defines the variation $\delta 
\bm{\theta}%
^{\mathsf{s}}$ of the 'quasi-rotation vector' (sec. \ref{secSpatialOmega}),
and $\delta 
\bm{\rho}%
^{\mathsf{s}}=\delta \mathbf{r}-\delta \widetilde{%
\bm{\theta}%
}\mathbf{r}$ is the variation of the position vector of that point attached
to the body which momentarily coincides with the origin of the IFR. Both
vectors are represented in the IFR. Noting that $d\delta \mathbf{C}-\delta d%
\mathbf{C}\equiv \mathbf{0}$, yields (omitting the subscript \textsf{s}) 
\begin{eqnarray}
d\delta \widehat{%
\bm{\pi}%
}-\delta d\widehat{%
\bm{\pi}%
} &=&d\left( \delta \mathbf{CC}^{-1}\right) -\delta \left( d\mathbf{CC}%
^{-1}\right)  \notag \\
&=&d\delta \mathbf{CC}^{-1}-\delta \mathbf{CC}^{-1}d\mathbf{CC}^{-1}-\delta
dC\mathbf{C}^{-1}+d\mathbf{CC}^{-1}\delta \mathbf{CC}^{-1}  \notag \\
&=&\left( d\delta \mathbf{C}-\delta d\mathbf{C}\right) \mathbf{C}^{-1}+d%
\widehat{%
\bm{\pi}%
}\delta \widehat{%
\bm{\pi}%
}-\delta \widehat{%
\bm{\pi}%
}d\widehat{%
\bm{\pi}%
}  \notag \\
&=&[d\widehat{%
\bm{\pi}%
},\delta \widehat{%
\bm{\pi}%
}].
\end{eqnarray}%
Inserting this in the transitory relation (\ref{transpos}), and using the
vector form (\ref{SE3Lie}) of the Lie bracket leads to%
\begin{equation}
\frac{d}{dt}\left( d\delta 
\bm{\pi}%
-\delta d%
\bm{\pi}%
\right) =\mathbf{ad}_{\mathbf{V}}\delta 
\bm{\pi}%
.
\end{equation}%
The coordinate expression (\ref{SE3Lie2}) of the Lie bracket yields, with $d%
\bm{\pi}%
/dt=\mathbf{V}$,%
\begin{equation}
\frac{d}{dt}\left( d\delta \pi _{k}-\delta d\pi _{k}\right) =-c_{kij}\delta
\pi _{i}V_{j}.  \label{spatial1}
\end{equation}%
Comparing (\ref{spatial1}) with (\ref{transpos}) shows that the Hamel
coefficients are 
\begin{equation}
\gamma _{ij}^{k}=-c_{kij},  \label{gammaVs}
\end{equation}%
which are identical to those in (\ref{transBody2}) when using the the
body-fixed representation of twists, but with opposite sign.

\subsubsection{Hybrid representation of rigid body twists}

A representation of twists frequently found in the literature, in particular
for computational multibody dynamics, is the so-called hybrid representation 
\cite{Murray}. The twist vector is denoted with $\mathbf{V}^{\mathrm{h}%
}=\left( 
\bm{\omega}%
^{\mathsf{s}},\dot{\mathbf{r}}\right) $ as defined in (\ref{Vh}). The name
indicates that it involves the velocity of the origin of the body-fixed
frame but the vectors are resolved in the IFR. The underlying geometric
model is the \emph{direct} product $SO\left( 3\right) \times {\mathbb{R}}%
^{3} $ with the group operation $C_{1}\cdot C_{2}=\left( \mathbf{R}_{1}%
\mathbf{R}_{2},\mathbf{r}_{1}+\mathbf{r}_{2}\right) $. The latter is clearly
not describing rigid body motions. With $\dot{\mathbf{r}}$ being 'holonomic
velocities' the Hamel coefficients are readily deduced from those of the
spatial angular velocity (\ref{gammaWs}) as%
\begin{equation}
\gamma _{ij}^{k}=\left\{ 
\begin{array}{cc}
-\varepsilon _{kij,} & i,j,k=1,2,3 \\ 
0, & i,j,k=4,5,6.%
\end{array}%
\right.  \label{HamelCoeffHybrid}
\end{equation}

\section{Boltzmann-Hamel Equations for the Rigid Body%
\label{secBHEFree}%
}

\subsection{Kinetic Energy of a Spatially Moving Body}

An arbitrarily located body-fixed reference frame $\mathcal{F}_{\mathsf{b}}$
is used to express the kinetic energy. Denote with $%
\bm{\omega}%
$ the angular velocity of $\mathcal{F}_{\mathsf{b}}$, and with $\mathbf{v}$
the velocity of its origin, both measured in IFR. Let $\mathbf{d}$ be the
position vector from the origin of $\mathcal{F}_{\mathsf{b}}$ to the COM,
and denote with $%
\bm{\Theta}%
^{\mathsf{c}}$ the inertia tensor w.r.t. the COM. The latter and all vectors
are resolved in IFR $\mathcal{F}_{0}$. The kinetic energy is then%
\begin{eqnarray}
T\left( 
\bm{\omega}%
,\mathbf{v}\right) &=&\frac{1}{2}%
\bm{\omega}%
^{T}%
\bm{\Theta}%
\bm{\omega}%
+\frac{m}{2}\mathbf{v}^{T}\mathbf{v}+m%
\bm{\omega}%
^{T}\widetilde{\mathbf{d}}\mathbf{v}=\frac{1}{2}\left( 
\bm{\omega}%
^{T},\mathbf{v}^{T}\right) \left( 
\begin{array}{cc}
\bm{\Theta}
& m\widetilde{\mathbf{d}} \\ 
-m\widetilde{\mathbf{d}}\ \ \  & m\mathbf{I}%
\end{array}%
\right) \left( 
\begin{array}{c}
\bm{\omega}
\\ 
\mathbf{v}%
\end{array}%
\right) \\
&=&\frac{1}{2}\omega _{i}\Theta _{ij}\omega _{j}+\frac{m}{2}%
v_{i}v_{i}+m\varepsilon _{ijk}\omega _{i}d_{j}v_{k}  \notag
\end{eqnarray}%
where $%
\bm{\Theta}%
:=%
\bm{\Theta}%
^{\mathsf{c}}-m\widetilde{\mathbf{d}}\widetilde{\mathbf{d}}$ is the inertia
tensor w.r.t. to the origin of the general body-fixed frame $\mathcal{F}_{%
\mathsf{b}}$ and $\mathbf{d}$ is the position vector from $\mathcal{F}_{%
\mathsf{b}}$ to the COM (parallel axis theorem). This is the kinetic energy
in terms of the rigid body twist measured at a general point and resolved in
the IFR. It is important that the kinetic energy is deduced from the COM
motion regardless of which point is used to measure the body twist.

\subsection{Newton-Euler equations -- Body-fixed representation of twists}

\subsubsection{Kinetic Energy}

Denote with $\mathbf{V}^{\mathsf{c}}=(%
\bm{\omega}%
^{\mathsf{c}},\mathbf{v}^{\mathsf{c}})$ the body-fixed twist of a frame $%
\mathcal{F}_{\mathsf{c}}$ at the COM ($%
\bm{\omega}%
^{\mathsf{c}}$ is the angular velocity of the body, and $\mathbf{v}^{\mathsf{%
c}}$ is the velocity of the origin of $\mathcal{F}_{\mathsf{c}}$, both
resolved in $\mathcal{F}_{\mathsf{c}}$), and denote with $%
\bm{\Theta}%
^{\mathsf{c}}$ the inertia tensor w.r.t. $\mathcal{F}_{\mathsf{c}}$. With
the mass $m$, the kinetic energy is $T=\frac{1}{2}%
\bm{\omega}%
^{\mathsf{c}T}%
\bm{\Theta}%
^{\mathsf{c}}%
\bm{\omega}%
^{\mathsf{c}}+\frac{m}{2}\mathbf{v}^{\mathsf{c}}{}^{T}\mathbf{v}^{\mathsf{c}%
} $. Now let $\mathcal{F}_{\mathsf{b}}$ be a general body-fixed frame, with
body-fixed twist $\mathbf{V}^{\mathsf{b}}=(%
\bm{\omega}%
^{\mathsf{b}},\mathbf{v}^{\mathsf{b}})$, related to $\mathcal{F}_{\mathsf{c}%
} $ by the rotation matrix $\mathbf{R}^{\mathsf{bc}}$ and displacement
vector $^{\mathsf{b}}\mathbf{d}$ from the origin of $\mathcal{F}_{\mathsf{b}}
$ to that of $\mathcal{F}_{\mathsf{c}}$ resolved in $\mathcal{F}_{\mathsf{b}}
$. Then $%
\bm{\omega}%
^{\mathsf{c}}=(\mathbf{R}^{\mathsf{bc}})^{T}%
\bm{\omega}%
^{\mathsf{b}}$ and $\mathbf{v}^{\mathsf{c}}=(\mathbf{R}^{\mathsf{bc}})^{T}(%
\mathbf{v}^{\mathsf{b}}-\widetilde{%
\bm{\omega}%
}{^{\mathsf{b}}}{^{\mathsf{b}}\mathbf{d}})$. For simplicity, the superscript 
\textsf{b} in $%
\bm{\omega}%
^{\mathsf{b}}$ and $\mathbf{v}^{\mathsf{b}}$ will be omitted in the
following. The kinetic energy in terms of $\mathbf{V}^{\mathsf{b}}$ is%
\begin{eqnarray}
T\left( 
\bm{\omega}%
,\mathbf{v}\right) &=&\frac{1}{2}\left( 
\bm{\omega}%
^{T},\mathbf{v}^{T}\right) \left( 
\begin{array}{cc}
\bm{\Theta}%
^{\mathsf{b}} & m\widetilde{\mathbf{d}} \\ 
-m\widetilde{\mathbf{d}}\ \ \  & m\mathbf{I}%
\end{array}%
\right) \left( 
\begin{array}{c}
\bm{\omega}
\\ 
\mathbf{v}%
\end{array}%
\right) =\frac{1}{2}\mathbf{V}^{T}\mathbf{M}^{\mathsf{b}}\mathbf{V}  \notag
\\
&=&\frac{1}{2}%
\bm{\omega}%
^{T}%
\bm{\Theta}%
^{\mathsf{b}}%
\bm{\omega}%
+\frac{m}{2}\mathbf{v}^{T}\mathbf{v}+m%
\bm{\omega}%
^{T}\widetilde{\mathbf{d}}%
\red%
\mathbf{v}%
\color{black}
\notag \\
&=&\frac{1}{2}\omega _{i}\Theta _{ij}^{\mathsf{b}}\omega _{j}+\frac{m}{2}%
v_{i}v_{i}+m\varepsilon _{ijk}\omega _{i}d_{j}v_{k}  \label{TkinBody}
\end{eqnarray}%
where $\mathbf{M}^{\mathsf{b}}$ is the constant mass matrix expressed in
body-fixed frame with $%
\bm{\Theta}%
^{\mathsf{b}}=\mathbf{R}^{\mathsf{bc}}%
\bm{\Theta}%
^{\mathsf{c}}(\mathbf{R}^{\mathsf{bc}})^{T}-m\widetilde{\mathbf{d}}%
\widetilde{\mathbf{d}}$. The kinetic energy (\ref{TkinBody}) is invariant
under change of body-fixed reference frame, i.e. left-invariant under action
of $SE\left( 3\right) $.

\subsubsection{Boltzmann-Hamel equations}

For the body-fixed representation of twists, the kinetic energy (\ref%
{TkinBody}) does not depend on the pose of the body. Using the Hamel
coefficients (\ref{gammaVb}) yields to the BH-equations (\ref{BHEqu2})
(omitting superscript $\mathsf{b}$)%
\begin{eqnarray}
\frac{d}{dt}\frac{\partial T}{\partial \omega _{i}}+\frac{\partial T}{%
\partial \omega _{k}}\varepsilon _{kij}\omega _{j}+\frac{\partial T}{%
\partial v_{k}}\varepsilon _{kij}v_{j} &=&\tau _{i},\ a=i=1,2,3
\label{BHbody11} \\
\frac{d}{dt}\frac{\partial T}{\partial v_{i}}+\frac{\partial T}{\partial
v_{k}}\varepsilon _{kij}\omega _{j} &=&f_{i},\ \ a=i+3=4,5,6
\label{BHbody21}
\end{eqnarray}%
where $\tau _{i}$ and and $f_{i}$ are the components of the applied torque
and force expressed in the reference frame $\mathcal{F}_{\mathsf{b}}$.
Inserting the derivatives%
\begin{eqnarray}
\frac{\partial T}{\partial \omega _{i}} &=&\Theta _{ij}^{\mathsf{b}}\omega
_{j}+m\varepsilon _{ijk}d_{j}v_{k},\ \ \ \frac{d}{dt}\frac{\partial T}{%
\partial \omega _{i}}=\Theta _{ij}^{\mathsf{b}}\dot{\omega}_{j}+m\varepsilon
_{ijk}d_{j}\dot{v}_{k}  \notag \\
\frac{\partial T}{\partial v_{i}} &=&m(v_{i}+\varepsilon _{ijk}\omega
_{j}d_{k}),\ \ \ \frac{d}{dt}\frac{\partial T}{\partial v_{i}}=m(\dot{v}%
_{i}+\varepsilon _{ijk}\dot{\omega}_{j}d_{k})  \label{bodyTmp}
\end{eqnarray}%
in (\ref{BHbody11}) yields 
\begin{equation}
\Theta _{ij}\dot{\omega}_{j}+m\varepsilon _{ijk}d_{j}\dot{v}_{k}+\varepsilon
_{ijk}\omega _{j}\Theta _{kl}\omega _{l}+m\varepsilon _{ijk}(\omega
_{j}\varepsilon _{krs}d_{r}v_{s}+v_{j}\varepsilon _{krs}\omega
_{r}d_{s})=\tau _{i}.  \label{BHbody12}
\end{equation}%
The last term can be simplified, invoking the Jacobi identity $\varepsilon
_{ijk}\varepsilon _{krs}+\varepsilon _{isk}\varepsilon _{kjr}=-\varepsilon
_{irk}\varepsilon _{ksj}$. With this, and introducing (\ref{bodyTmp}) in the
second equation (\ref{BHbody21}) yields%
\begin{eqnarray}
\Theta _{ij}^{\mathsf{b}}\dot{\omega}_{j}+\varepsilon _{ijk}\omega
_{j}\Theta _{kl}^{\mathsf{b}}\omega _{l}+m\varepsilon _{ijk}d_{j}(\dot{v}%
_{k}+\varepsilon _{krs}\omega _{s}v_{r}) &=&\tau _{i}  \notag \\
m(\dot{v}_{i}+\varepsilon _{ijk}\omega _{j}v_{k}+(\varepsilon _{ijs}\dot{%
\omega}_{j}+\varepsilon _{ijk}\varepsilon _{krs}\omega _{j}\omega
_{r})d_{s}) &=&f_{i}.  \label{BHBodyFixed}
\end{eqnarray}%
These are the NE-equations w.r.t. to an arbitrary body-fixed frame, which
read in vector form%
\begin{eqnarray}
\bm{\Theta}%
^{\mathsf{b}}\dot{%
\bm{\omega}%
}^{\mathsf{b}}+\widetilde{%
\bm{\omega}%
}{^{\mathsf{b}}}%
\bm{\Theta}%
^{\mathsf{b}}%
\bm{\omega}%
^{\mathsf{b}}\color{red}-m{^{\mathsf{b}}}\left( \dot{\mathbf{v}}^{\mathsf{b}%
}+\widetilde{%
\bm{\omega}%
}{^{\mathsf{b}}}\mathbf{v}{^{\mathsf{b}}}\right) \widetilde{\mathbf{d}}_{%
\text{bc}}\color{black}) &=&%
\bm{\tau}%
{^{\mathsf{b}}}  \notag \\
m%
\big%
(\dot{\mathbf{v}}{^{\mathsf{b}}}+\widetilde{%
\bm{\omega}%
}{^{\mathsf{b}}}\mathbf{v}{^{\mathsf{b}}}+(\dot{\widetilde{%
\bm{\omega}%
}}{^{\mathsf{b}}}+\widetilde{%
\bm{\omega}%
}{^{\mathsf{b}}}\widetilde{%
\bm{\omega}%
}{^{\mathsf{b}}})\mathbf{d}%
\big%
) &=&\mathbf{f}{^{\mathsf{b}}.}  \label{BHBodyFixedExplicit}
\end{eqnarray}%
They simplify if the body-fixed reference frame is located at the COM, i.e.
if $\mathbf{d}=\mathbf{0}$.

\subsubsection{BH-Equations as Euler-Poincar\'{e} Equations on $SE\left(
3\right) $}

The equations (\ref{BHBodyFixedExplicit}) can be written in the compact form 
\cite{MUBOScrews2}%
\begin{equation}
\mathbf{M}^{\mathsf{b}}\dot{\mathbf{V}}^{\mathsf{b}}-\mathbf{ad}_{\mathbf{V}%
^{\mathsf{b}}}^{T}\mathbf{M}^{\mathsf{b}}\mathbf{V}^{\mathsf{b}}=\mathbf{W}^{%
\mathsf{b}}  \label{NEBodyFixedMatrix}
\end{equation}%
where $\mathbf{W}^{\mathsf{b}}=(%
\bm{\tau}%
{^{\mathsf{b}},}\mathbf{f}{^{\mathsf{b}}})\in se^{\ast }\left( 3\right) $ is
the body-fixed representation of the wrench acting on the body. These are
the Euler-Poincar\'{e} equations (\ref{Poincare3}) for the left-invariant
kinetic energy (\ref{TkinBody}) (invariant w.r.t. choice of IFR). They can
be written as%
\begin{equation}
\dot{%
\bm{\Pi}%
}^{\mathsf{b}}-\mathbf{ad}_{\mathbf{V}^{\mathsf{b}}}^{T}%
\bm{\Pi}%
^{\mathsf{b}}=\mathbf{W}^{\mathsf{b}}  \label{NE}
\end{equation}%
in terms of the body-fixed momentum $%
\bm{\Pi}%
^{\mathsf{b}}:=\mathbf{M}^{\mathsf{b}}\mathbf{V}^{\mathsf{b}}\in se^{\ast
}\left( 3\right) $, with which the kinetic energy is $T(\mathbf{V}^{\mathsf{b%
}}):=\frac{1}{2}%
\bm{\Pi}%
^{\mathsf{b}}\mathbf{V}^{\mathsf{b}}$ .

\subsection{Newton-Euler equations -- Spatial representation of twists}

\subsubsection{Kinetic Energy}

The peculiarity of the spatial twist (\ref{SpatialTwist}) is that $\mathbf{v}%
^{\mathsf{s}}=\dot{\mathbf{r}}-\widetilde{\mathbf{r}}^{\mathsf{s}}%
\bm{\omega}%
^{\mathsf{s}}$ is the velocity of the point of the body which is momentarily
coinciding with the origin of the IFR (recall that $\mathbf{r}$ is the
position vector of the body-fixed frame pressed in IFR). Denote with $%
\mathbf{d}$ the vector from the body-fixed frame to the COM. The velocity of
the COM, expressed in IFR, is then $\mathbf{v}^{\mathsf{s}}+(\widetilde{%
\mathbf{d}}+\widetilde{\mathbf{r}})\widetilde{%
\bm{\omega}%
}^{\mathsf{s}}$, and the expression for the kinetic energy in terms of
spatial twists defined in (\ref{SpatialTwist}) is%
\begin{eqnarray}
T\left( 
\bm{\varphi}%
,\mathbf{r},%
\bm{\omega}%
^{\mathsf{s}},\mathbf{v}^{\mathsf{s}}\right) &=&\frac{1}{2}\left( 
\bm{\omega}%
^{\mathsf{s}T},\mathbf{v}^{\mathsf{s}T}\right) \left( 
\begin{array}{cc}
\bm{\Theta}%
^{\mathsf{s}} & m(\widetilde{\mathbf{d}}+\widetilde{\mathbf{r}}) \\ 
-m(\widetilde{\mathbf{d}}+\widetilde{\mathbf{r}})\ \ \  & m\mathbf{I}%
\end{array}%
\right) \left( 
\begin{array}{c}
\bm{\omega}%
^{\mathsf{s}} \\ 
\mathbf{v}^{\mathsf{s}}%
\end{array}%
\right) =\frac{1}{2}\mathbf{V}^{\mathsf{s}T}\mathbf{M}^{\mathsf{s}}\mathbf{V}%
^{\mathsf{s}}  \notag \\
&=&\frac{1}{2}\omega _{i}^{\mathsf{s}}\Theta _{ij}^{\mathsf{s}}\omega _{j}^{%
\mathsf{s}}+\frac{m}{2}v_{i}^{\mathsf{s}}v_{i}^{\mathsf{s}}+m\varepsilon
_{ijk}\omega _{i}^{\mathsf{s}}\left( d_{j}+r_{j}\right) v_{k}^{\mathsf{s}}
\label{Tspatial}
\end{eqnarray}%
where $%
\bm{\Theta}%
^{\mathsf{s}}={^{\mathsf{s}}%
\bm{\Theta}%
^{\mathsf{c}}}-m(\widetilde{\mathbf{d}}+\widetilde{\mathbf{r}})$ is the
inertia tensor w.r.t. the body-fixed point that coincides with the IFR
origin resolved in IFR. Therein, ${^{\mathsf{s}}%
\bm{\Theta}%
^{\mathsf{c}}=\mathbf{R}%
\bm{\Theta}%
^{\mathsf{c}}}\mathbf{R}^{T}$ is the inertia tensor w.r.t. the COM, resolved
in the IFR, where ${\mathbf{R}}$ is the rotation matrix in (\ref{C})
transforming from body-fixed frame $\mathcal{F}_{\mathsf{b}}$ to IFR $%
\mathcal{F}_{0}$. Its components are 
\begin{equation}
\Theta _{ij}^{\mathsf{s}}\left( 
\bm{\varphi}%
\right) ={^{\mathsf{s}}\Theta _{ij}^{\mathsf{c}}}\left( 
\bm{\varphi}%
\right) -m\varepsilon _{ikl}\varepsilon _{lrj}\left( d_{k}+r_{k}\right)
\left( d_{r}+r_{r}\right) .  \label{Thetac}
\end{equation}%
Instead of writing $C\in SE\left( 3\right) $, local coordinates $\left(
q^{a}\right) =\left( \varphi ^{i},r_{i}\right) $ are used in (\ref{Tspatial}%
) to express the rigid body pose. The superscript \textsf{s} in $%
\bm{\omega}%
^{\mathsf{s}}$ and $\mathbf{v}^{\mathsf{s}}$ will be omitted in the
following. All vectors are resolved in the IFR. The kinetic energy (\ref%
{Tspatial}) is invariant under change of IFR, i.e. right-invariant under
action of $SE\left( 3\right) $.

\subsubsection{Boltzmann-Hamel equations%
\label{secBHSpatial}%
}

With the Hamel coefficients (\ref{gammaVs}) and the kinetic energy, the
BH-equations (\ref{BHEqu2}) are (omitting the superscript $\mathsf{s}$ of $%
v_{i}^{\mathsf{s}}$ and $\omega _{i}^{\mathsf{s}}$)%
\begin{eqnarray}
\frac{d}{dt}\frac{\partial T}{\partial \omega _{i}}-\frac{\partial T}{%
\partial \varphi ^{j}}\frac{\partial \dot{\varphi}^{j}}{\partial \omega _{i}}%
-\frac{\partial T}{\partial r_{j}}\frac{\partial \dot{r}_{j}}{\partial
\omega _{i}}-\frac{\partial T}{\partial \omega _{k}}\varepsilon _{kij}\omega
_{j}-\frac{\partial T}{\partial v_{k}}\varepsilon _{kij}v_{j} &=&\tau _{i},\
a=i=1,2,3  \label{BHspatial11} \\
\frac{d}{dt}\frac{\partial T}{\partial v_{i}}-\frac{\partial T}{\partial
r_{i}}-\frac{\partial T}{\partial v_{k}}\varepsilon _{kij}\omega _{j}
&=&f_{i},\ \ a=i+3=4,5,6.  \label{BHspatial12}
\end{eqnarray}%
For the subsequent derivation, the following relations will be necessary,
which follow from the definition of the spatial twist (\ref{SpatialTwist})
and from (\ref{Aij}):%
\begin{equation}
\dot{r}_{i}=v_{i}-\varepsilon _{ijk}r_{j}\omega _{k},\ \frac{\partial \dot{r}%
_{j}}{\partial \omega _{i}}=-\varepsilon _{jki}r_{k},\ \frac{\partial \dot{%
\varphi}_{j}}{\partial \omega _{i}}=\Psi _{i}^{j}.  \label{temp3}
\end{equation}%
Therewith the first set of equations (\ref{BHspatial11}), describing the
angular dynamics, is%
\begin{equation}
\frac{d}{dt}\frac{\partial T}{\partial \omega _{i}}-\frac{\partial T}{%
\partial \varphi ^{j}}\Psi _{i}^{j}+\frac{\partial T}{\partial r_{j}}%
\varepsilon _{jki}r_{k}-\frac{\partial T}{\partial \omega _{k}}\varepsilon
_{kij}\omega _{j}-\frac{\partial T}{\partial v_{k}}\varepsilon
_{kij}v_{j}=\tau _{i}.  \label{temp1}
\end{equation}%
From the kinetic energy (\ref{Tspatial}) follows immediately%
\begin{equation}
\frac{\partial T}{\partial \omega _{i}}=\Theta _{ij}^{\mathsf{s}}\omega
_{j}+m\varepsilon _{ijk}\left( d_{j}+r_{j}\right) v_{k}  \label{dTdw}
\end{equation}%
and inserting (\ref{Thetac}) yields%
\begin{equation*}
\frac{d}{dt}\frac{\partial T}{\partial \omega _{i}}=\Theta _{ij}^{\mathsf{s}}%
\dot{\omega}_{j}+(^{\mathsf{s}}\dot{\Theta}_{ij}^{\mathsf{c}}-m\varepsilon
_{ikl}\varepsilon _{lrj}\left( \left( d_{k}+r_{k}\right) \left( \dot{d}_{r}+%
\dot{r}_{r}\right) +\left( \dot{d}_{k}+\dot{r}_{k}\right) \left(
d_{r}+r_{r}\right) \right) )\omega _{j}+m\varepsilon _{ijk}\left( \left(
d_{j}+r_{j}\right) \dot{v}_{k}+(\dot{d}_{j}+\dot{r}_{j})v_{k}\right) .
\end{equation*}%
Introducing the relations $\dot{d}_{j}=\varepsilon _{jlr}\omega _{l}d_{r}$
and $^{\mathsf{s}}\dot{\Theta}_{ij}^{\mathsf{c}}=(\varepsilon _{ikl}{^{%
\mathsf{s}}\Theta _{lj}^{\mathsf{c}}}-{^{\mathsf{s}}\Theta _{il}^{\mathsf{c}}%
}\varepsilon _{klj})\omega _{k}$ along with (\ref{temp3}) leads to%
\begin{eqnarray}
\frac{d}{dt}\frac{\partial T}{\partial \omega _{i}} &=&\Theta _{ij}^{\mathsf{%
s}}\dot{\omega}_{j}+\varepsilon _{ikl}\omega _{k}{^{\mathsf{s}}\Theta _{lj}^{%
\mathsf{c}}}\omega _{j}+m\varepsilon _{ijk}\left( d_{j}+r_{j}\right) \dot{v}%
_{k}-m\varepsilon _{ikj}\varepsilon _{jlr}\left( d_{k}+r_{k}\right)
v_{l}\omega _{r}  \notag \\
&&+m\varepsilon _{ikl}\left( \varepsilon _{ljr}\varepsilon
_{rsp}(d_{k}+r_{k})\omega _{j}\omega _{s}\left( d_{p}+r_{p}\right)
-\varepsilon _{ksp}\varepsilon _{lrj}\omega _{s}(d_{p}+r_{p})\left(
d_{r}+r_{r}\right) \omega _{j}\right)  \label{temp2}
\end{eqnarray}%
The term in the second line of (\ref{temp2}) can be summarized as $%
-m\varepsilon _{ikl}\varepsilon _{lsp}\varepsilon _{prj}\omega
_{k}(d_{s}+r_{s})\omega _{j}$, and (\ref{temp2}) simplifies to%
\begin{equation}
\frac{d}{dt}\frac{\partial T}{\partial \omega _{i}}=\Theta _{ij}^{\mathsf{s}}%
\dot{\omega}_{j}+\varepsilon _{ikl}\omega _{k}\Theta _{lj}^{\mathsf{s}%
}\omega _{j}+m\varepsilon _{ijk}\left( d_{j}+r_{j}\right) \left( \dot{v}%
_{k}+\varepsilon _{jlr}\omega _{l}v_{r}\right) .  \label{dtdTdw}
\end{equation}%
The partial derivative of the kinetic energy (\ref{Tspatial}) w.r.t. $%
\varphi $ follows with (\ref{Thetac}) as%
\begin{eqnarray}
\frac{\partial T}{\partial \varphi ^{i}} &=&\frac{1}{2}\omega _{k}\frac{%
\partial {^{\mathsf{s}}\Theta _{kj}^{\mathsf{c}}}}{\partial \varphi ^{i}}%
\omega _{j}+\frac{1}{2}m\varepsilon _{ljk}\frac{\partial d_{j}}{\partial
\varphi ^{i}}v_{k}\omega _{l}  \notag \\
&=&\frac{1}{2}\omega _{k}%
\Big%
(\frac{\partial {^{\mathsf{s}}\Theta _{kj}^{\mathsf{c}}}}{\partial \varphi
^{i}}-m\varepsilon _{ksl}\varepsilon _{lrj}%
\Big%
(\frac{\partial d_{s}}{\partial \varphi ^{i}}\left( d_{r}+r_{r}\right)
+\left( d_{s}+r_{s}\right) \frac{\partial d_{r}}{\partial \varphi ^{i}}%
\Big%
)%
\Big%
)\omega _{j}+m\omega _{k}\varepsilon _{kjl}\frac{\partial d_{j}}{\partial
\varphi ^{i}}v_{l}.
\end{eqnarray}%
Introducing $\frac{\partial d_{j}}{\partial \varphi ^{i}}=\varepsilon
_{jrs}\Omega _{ri}d_{s}$ and $\frac{\partial {^{\mathsf{s}}\Theta _{kj}^{%
\mathsf{c}}}}{\partial \varphi ^{i}}=\varepsilon _{kls}\Omega _{li}{^{%
\mathsf{s}}\Theta _{sj}^{\mathsf{c}}}-{^{\mathsf{s}}\Theta _{ks}^{\mathsf{c}}%
}\varepsilon _{slj}\Omega _{li}$, some straightforward manipulation yields 
\begin{equation}
\frac{\partial T}{\partial \varphi ^{i}}=-\Omega _{ji}\varepsilon
_{jsl}\omega _{s}{^{\mathsf{s}}\Theta _{lk}^{\mathsf{c}}}\omega _{k}+m\omega
_{k}\varepsilon _{kjl}\varepsilon _{jrs}\Omega _{ri}d_{s}v_{l}-m\omega
_{k}\varepsilon _{ksl}\varepsilon _{spn}\Omega _{pi}d_{n}\varepsilon
_{lrj}\left( d_{r}+r_{r}\right) \omega _{j}.  \label{dTdphi}
\end{equation}%
The derivative w.r.t. $r_{k}$ can be simplified with $\varepsilon
_{ikl}\varepsilon _{lrj}+\varepsilon _{irl}\varepsilon _{lkj}=2\varepsilon
_{ikl}\varepsilon _{lrj}+\varepsilon _{lrj}\varepsilon _{lkr}$ to yield%
\begin{eqnarray}
\frac{\partial T}{\partial r_{k}} &=&m\varepsilon _{ikl}\omega _{i}v_{j}-%
\frac{m}{2}\omega _{i}\left( \varepsilon _{ikl}\varepsilon _{lrj}\left(
d_{r}+r_{r}\right) +\varepsilon _{irl}\varepsilon _{lkj}\left(
d_{r}+r_{r}\right) \right) \omega _{j}  \notag \\
&=&m\varepsilon _{ikl}\omega _{i}v_{j}-m\omega _{i}\varepsilon
_{ikl}\varepsilon _{lrj}\left( d_{r}+r_{r}\right) \omega _{j}.  \label{dTdr}
\end{eqnarray}%
Finally, it is readily found that 
\begin{equation}
\frac{\partial T}{\partial v_{k}}=m\left( \dot{v}_{i}+\varepsilon _{ijk}\dot{%
\omega}_{j}\left( d_{k}+r_{k}\right) +\varepsilon _{ijk}\omega _{j}\left(
v_{k}+\varepsilon _{klr}\omega _{l}\left( d_{r}+r_{r}\right) \right) \right)
.  \label{dTdv}
\end{equation}%
Inserting (\ref{dTdw}), (\ref{dTdphi}), (\ref{dTdr}), and (\ref{dTdv}) into
the BH-equation (\ref{BHspatial11}), noting that $\Omega _{ik}\Psi
_{j}^{k}=\delta _{ij}$, and summarizing terms, yields%
\begin{eqnarray}
\tau _{i} &=&\frac{d}{dt}\frac{\partial T}{\partial \omega _{i}}  \notag \\
(\ast ) &&+\omega _{s}\left( \varepsilon _{isl}{^{\mathsf{s}}\Theta _{lk}^{%
\mathsf{c}}}\omega _{k}-\varepsilon _{isk}{^{\mathsf{s}}\Theta _{kl}^{%
\mathsf{c}}}\omega _{l}\right) -mv_{k}\varepsilon _{kij}v_{j}  \notag \\
(\ast \ast ) &&+m\omega _{k}\varepsilon _{ksl}\varepsilon
_{sin}d_{n}\varepsilon _{lrj}\left( d_{r}+r_{r}\right) \omega _{j}+m\omega
_{k}\varepsilon _{sjl}\varepsilon _{jip}r_{p}\varepsilon _{lrk}\left(
d_{r}+r_{r}\right) \omega _{k}+m\omega _{j}\varepsilon _{ijk}\varepsilon
_{ksl}\left( d_{s}+r_{s}\right) \varepsilon _{lrp}\left( d_{r}+r_{r}\right)
\omega _{p}  \notag \\
(\ast \ast \ast ) &&-m\omega _{j}\varepsilon _{ijk}\varepsilon _{krs}\left(
d_{r}+r_{r}\right) v_{s}-mv_{j}\varepsilon _{ijk}\varepsilon _{krs}\omega
_{r}\left( d_{s}+r_{s}\right)  \notag \\
(\ast \ast \ast \ast ) &&+md_{s}\varepsilon _{isj}\varepsilon _{jkl}\omega
_{k}v_{l}+mr_{k}\varepsilon _{ikj}\varepsilon _{jrs}\omega _{r}v_{s}.
\label{temp4}
\end{eqnarray}%
The expression (\ref{dtdTdw}) was not inserted on purpose. The terms $(\ast
) $ and $(\ast \ast )$ clearly vanish. Due to the Jacobi identity, $(\ast
\ast \ast )$ is equal to $m\varepsilon _{ijk}\left( d_{j}+r_{j}\right)
\varepsilon _{krs}v_{r}\omega _{s}$. Summarizing the term $(\ast \ast \ast
\ast )$ as $m\varepsilon _{ijk}\left( d_{j}+r_{j}\right) \varepsilon
_{krs}\omega _{r}v_{s}$ shows that $(\ast \ast \ast )+(\ast \ast \ast \ast
)=0$. The equation (\ref{BHspatial11}) thus attains the simple form%
\begin{eqnarray}
\tau _{i} &=&\frac{d}{dt}\frac{\partial T}{\partial \omega _{i}}  \notag \\
&=&\Theta _{ij}^{\mathsf{s}}\dot{\omega}_{j}+\varepsilon _{ikl}\omega
_{k}\Theta _{lj}^{\mathsf{s}}\omega _{j}+m\varepsilon _{ijk}\left(
d_{j}+r_{j}\right) \left( \dot{v}_{k}+\varepsilon _{jlr}\omega
_{l}v_{r}\right) ,\ \ i=1,2,3.  \label{BHspatial21}
\end{eqnarray}

The second set of equations (\ref{BHspatial12}) yields, with (\ref{dTdr})
and (\ref{dTdv}), and after simplification%
\begin{equation}
f_{i}=m\left( \dot{v}_{i}-\varepsilon _{ijk}\left( d_{j}+r_{j}\right) \dot{%
\omega}_{k}+\varepsilon _{ijk}\omega _{j}\left( v_{k}+\varepsilon
_{klr}\omega _{l}\left( d_{r}+r_{r}\right) \right) \right) ,\ \ i=1,2,3.
\label{BHspatial22}
\end{equation}%
The equations (\ref{BHspatial21}) and (\ref{BHspatial22}) are the
Newton-Euler equations governing the rigid body dynamics in terms of the
spatial twist. They are in vector form%
\begin{eqnarray}
\bm{\Theta}%
^{\mathsf{s}}\dot{%
\bm{\omega}%
}^{\mathsf{s}}+\widetilde{%
\bm{\omega}%
}{^{\mathsf{s}}}%
\bm{\Theta}%
^{\mathsf{s}}%
\bm{\omega}%
^{\mathsf{s}}+m(\widetilde{\mathbf{d}}+\widetilde{\mathbf{r}})(\dot{\mathbf{v%
}}{^{\mathsf{s}}+}\widetilde{%
\bm{\omega}%
}{^{\mathsf{s}}}\mathbf{v}{^{\mathsf{s}}}) &=&%
\bm{\tau}%
{^{\mathsf{s}}}  \label{NESpatial1} \\
m%
\big%
(\dot{\mathbf{v}}{^{\mathsf{s}}}+\widetilde{%
\bm{\omega}%
}{^{\mathsf{s}}}\mathbf{v}{^{\mathsf{s}}}+(\dot{\widetilde{%
\bm{\omega}%
}}{^{\mathsf{s}}}+\widetilde{%
\bm{\omega}%
}{^{\mathsf{s}}}\widetilde{%
\bm{\omega}%
}{^{\mathsf{s}}})\left( \mathbf{d}+\mathbf{r}\right) 
\big%
) &=&\mathbf{f}{^{\mathsf{s}}}  \label{NESpatial2}
\end{eqnarray}%
These equations cannot be simplified, as (\ref{BHBodyFixedExplicit}) can be
by setting $\mathbf{d}=\mathbf{0}$, since the COM cannot be used as
reference point.

\subsubsection{BH-Equations as Euler-Lagrange Equations on $SE\left(
3\right) $}

Also the spatial NE equations (\ref{BHspatial21},\ref{BHspatial22}) can be
written in compact vector form%
\begin{equation}
\mathbf{M}^{\mathsf{s}}\dot{\mathbf{V}}^{\mathsf{s}}-\mathbf{ad}_{\mathbf{V}%
^{\mathsf{s}}}^{T}\mathbf{M}^{\mathsf{s}}\mathbf{V}^{\mathsf{s}}=\mathbf{W}^{%
\mathsf{s}}  \label{NESpatial}
\end{equation}%
where $\mathbf{W}^{\mathsf{s}}=\left( 
\bm{\tau}%
^{\mathsf{s}},\mathbf{f}^{\mathsf{s}}\right) \in se^{\ast }\left( 3\right) $
is the applied wrench in spatial representation. These are the
Euler-Lagrange equations (\ref{EL}) for the right-invariant kinetic energy.
Although (\ref{NESpatial}) seem formally identical to (\ref{Poincare3}), and
to (\ref{NEBodyFixedMatrix}) in particular, they are not the Euler-Poincar%
\'{e} equations (which would require a positive sign in (\ref{NESpatial}).
The kinetic energy in terms of spatial twists is right invariant
(independent from the choice of body-fixed frame), but it depends on the
orientation and thus on the pose of the body: $T=T\left( C,\mathbf{V}^{%
\mathsf{s}}\right) $.

\begin{remark}
The equations (\ref{NESpatial1}) and (\ref{NESpatial2}) can be readily
derived from the conservation of the spatial momentum $\bm{\Pi}^{\mathsf{s}}=%
\mathbf{M}^{\mathsf{s}}\mathbf{V}^{\mathsf{s}}$, using $\dot{\mathbf{M}}^{%
\mathsf{s}}=-\mathbf{M}^{\mathsf{s}}\mathbf{ad}_{\mathbf{V}^{\mathsf{s}}}-%
\mathbf{ad}_{\mathbf{V}^{\mathsf{s}}}^{T}\mathbf{M}^{\mathsf{s}}$, as%
\begin{equation}
\dot{\bm{\Pi}}^{\mathsf{s}}=\mathbf{W}^{\mathsf{s}}.  \label{SpatialConserve}
\end{equation}%
Beside its simplicity, the spatial form (\ref{SpatialConserve}) is also
computationally advantageous. An important advantage is that momentum
conserving numerical integration schemes can be easily derived, including
Lie group integration schemes. This was discussed in \cite%
{Borri2001b,Borri2003,Bottasso1998}. In this context, the formulation in
terms of spatial twists was called fixed-pole formulation.
\end{remark}

\subsection{Newton-Euler equations -- Hybrid representation of twists}

\subsubsection{Kinetic Energy}

The kinetic energy in terms of the hybrid twist (\ref{Vh}) is

\begin{eqnarray}
T\left( 
\bm{\varphi}%
,%
\bm{\omega}%
^{\mathsf{s}},\dot{\mathbf{r}}\right) &=&\frac{1}{2}\left( 
\bm{\omega}%
^{\mathsf{s}T},\dot{\mathbf{r}}^{T}\right) \left( 
\begin{array}{cc}
{^{\mathsf{s}}%
\bm{\Theta}%
^{\mathsf{b}}} & m\widetilde{\mathbf{d}} \\ 
-m\widetilde{\mathbf{d}}\ \ \  & m\mathbf{I}%
\end{array}%
\right) \left( 
\begin{array}{c}
\bm{\omega}%
^{\mathsf{s}} \\ 
\dot{\mathbf{r}}%
\end{array}%
\right) =\frac{1}{2}\mathbf{V}^{\mathsf{h}T}\mathbf{M}^{\mathsf{h}}\mathbf{V}%
^{\mathsf{h}}  \label{EkinHybrid} \\
&=&\frac{1}{2}%
\bm{\omega}%
^{\mathsf{s}T}{^{\mathsf{s}}%
\bm{\Theta}%
^{\mathsf{b}}}%
\bm{\omega}%
^{\mathsf{s}}+\frac{m}{2}\dot{\mathbf{r}}^{T}\dot{\mathbf{r}}+m%
\bm{\omega}%
^{\mathsf{s}}{}^{T}\widetilde{\mathbf{d}}\dot{\mathbf{r}}=\frac{1}{2}\omega
_{i}^{\mathsf{s}}{\Theta _{ij}^{\mathsf{b}}}\omega _{j}^{\mathsf{s}}+\frac{m%
}{2}\dot{r}_{i}\dot{r}_{i}+m\varepsilon _{ijk}\omega _{i}^{\mathsf{s}}d_{j}%
\dot{r}_{k}  \notag
\end{eqnarray}%
where ${^{\mathsf{s}}%
\bm{\Theta}%
^{\mathsf{b}}=\mathbf{R}%
\bm{\Theta}%
^{\mathsf{b}}}\mathbf{R}^{T}$ is the inertia tensor w.r.t. the (arbitrarily
located) body-fixed frame (superscript \textsf{b}) resolved in the IFR
(superscript \textsf{s}). All vectors are resolved in the IFR. The
superscripts of $%
\bm{\omega}%
^{\mathsf{s}}$ and ${^{\mathsf{s}}%
\bm{\Theta}%
^{\mathsf{b}}}$ will be omitted. The kinetic energy (\ref{EkinHybrid}) is
right-invariant under action of $SO\left( 3\right) \times {\mathbb{R}}^{3}$,
i.e. it is invariant w.r.t. rotations of IFR and displacements of its origin
taking place independently.

\subsubsection{Boltzmann-Hamel equations}

With the Hamel coefficients (\ref{HamelCoeffHybrid}) being zero for $%
a,b,c=4,5,6$, and with kinetic energy being independent from $\mathbf{r}$,
the BH-equations are%
\begin{eqnarray}
\frac{d}{dt}\frac{\partial T}{\partial \omega _{i}}-\frac{\partial T}{%
\partial \varphi ^{j}}\Psi _{i}^{j}+\frac{\partial T}{\partial \omega _{k}}%
\gamma _{ij}^{k}\omega _{j} &=&\tau _{i}^{\mathsf{s}},\ a=i=1,2,3
\label{BHEhybrid1} \\
\frac{d}{dt}\frac{\partial T}{\partial \dot{r}_{i}} &=&f_{i}^{\mathsf{h}},\
\ a=i+3=4,5,6  \label{BHEhybrid2}
\end{eqnarray}%
where $f_{i}^{\mathsf{h}}=R_{ij}f_{j}^{\mathsf{b}}$ are the coordinates of
the force applied at the origin of the body-fixed reference frame resolved
in the IFR, and $\tau _{i}^{\mathsf{s}}$ coordinates of the spatial torque
as in (\ref{BHspatial21}). From (\ref{EkinHybrid}) follows that $\frac{%
\partial T}{\partial \omega _{i}}={\Theta _{ij}}\omega _{j}+m\varepsilon
_{ijk}d_{j}\dot{r}_{k}$ and thus%
\begin{eqnarray}
\frac{d}{dt}\frac{\partial T}{\partial \omega _{i}} &=&\frac{d}{dt}\left(
\Theta _{ij}\omega _{j}\right) +m\varepsilon _{ijk}\left( d_{j}\ddot{r}_{k}+%
\dot{d}_{j}\dot{r}_{k}\right)  \notag \\
&=&\frac{d}{dt}\left( \Theta _{ij}\omega _{j}\right) +m\varepsilon
_{ijk}d_{j}\ddot{r}_{k}+m\varepsilon _{ijk}\varepsilon _{jrl}\omega _{r}d_{l}%
\dot{r}_{k}
\end{eqnarray}%
where $\dot{d}_{j}=\varepsilon _{jrl}\omega _{r}d_{l}$ has been used. Using $%
\frac{\partial d_{j}}{\partial \varphi _{i}}=\varepsilon _{jrs}\Omega
_{ri}d_{s}\dot{r}_{k}$ yields the partial derivatives%
\begin{eqnarray}
\frac{\partial T}{\partial \varphi _{i}} &=&\frac{1}{2}\omega _{k}\frac{%
\partial \Theta _{kj}}{\partial \varphi ^{i}}\omega _{j}+m\varepsilon
_{ljk}\varepsilon _{jrs}\omega _{l}\Omega _{ri}d_{s}\dot{r}_{k} \\
\frac{d}{dt}\frac{\partial T}{\partial \dot{r}_{i}} &=&m\ddot{r}%
_{i}+m\varepsilon _{ijk}(\dot{\omega}_{j}d_{k}+\omega _{j}\dot{d}_{k}) \\
&=&m(\ddot{r}_{i}+\varepsilon _{ijk}\dot{\omega}_{j}d_{k}+\varepsilon
_{ijk}\varepsilon _{krs}\omega _{j}\omega _{r}d_{s}).  \label{dTdrdot}
\end{eqnarray}%
The BH-equations (\ref{BHEhybrid1}) for $i=1,2,3$ are thus, with $\gamma
_{ij}^{k}=-\varepsilon _{kij}$ in (\ref{HamelCoeffHybrid}),%
\begin{eqnarray}
\tau _{i} &=&\frac{d}{dt}\left( \Theta _{ij}\omega _{j}\right) -\frac{1}{2}%
\omega _{k}\frac{\partial \Theta _{kr}}{\partial \varphi ^{j}}\omega
_{r}\Psi _{i}^{j}-\varepsilon _{kij}\Theta _{ks}\omega _{s}\omega _{j} 
\notag \\
&&+m\varepsilon _{ijk}d_{j}\ddot{r}_{k}  \label{BHhybrid1b} \\
&&+m\varepsilon _{ijk}\varepsilon _{jrl}\omega _{r}d_{l}\dot{r}%
_{k}-m\varepsilon _{lpk}\varepsilon _{prs}\omega _{l}\Omega _{rj}d_{s}\dot{r}%
_{k}\Psi _{i}^{j}-m\varepsilon _{kij}\varepsilon _{krs}d_{r}\dot{r}%
_{s}\omega _{j}.  \notag
\end{eqnarray}%
With $\Omega _{rj}\Psi _{i}^{j}=\delta _{ri}$, the last line in (\ref%
{BHhybrid1b}) can be rearranged as $-m(\varepsilon _{ijk}\varepsilon
_{krs}+\varepsilon _{isk}\varepsilon _{kjl}+\varepsilon _{irk}\varepsilon
_{ksj})\dot{r}_{j}\omega _{r}d_{s}\equiv 0$, and thus vanishes identically.
The right hand side of the first line in (\ref{BHhybrid1b}) is identical to
the left hand hand side of (\ref{BHRotSpatial2}), which yields the
Euler-equations (\ref{EulerSpatial}). Thus, the first set of BH-equations (%
\ref{BHEhybrid1}) are%
\begin{equation}
{^{\mathsf{s}}\Theta _{ij}^{\mathsf{b}}}\dot{\omega}_{j}^{\mathsf{s}%
}+\varepsilon _{ikl}\omega _{k}^{\mathsf{s}}{^{\mathsf{s}}\Theta _{lj}^{%
\mathsf{b}}}\omega _{j}^{\mathsf{s}}+m\varepsilon _{ijk}d_{j}\ddot{r}%
_{k}=\tau _{i}^{\mathsf{s}}.
\end{equation}%
The second set of BH-equations (\ref{BHRotSpatial2}) follows immediately
from (\ref{dTdrdot}) as 
\begin{equation}
m(\ddot{r}_{i}+\varepsilon _{ijk}\dot{\omega}_{j}^{\mathsf{s}%
}d_{k}+\varepsilon _{ijk}\varepsilon _{krs}\omega _{j}^{\mathsf{s}}\omega
_{r}^{\mathsf{s}}d_{s})=f_{i}^{\mathsf{h}}.
\end{equation}%
These are the Newton-Euler equations in terms of the hybrid twists, i.e.
spatial angular and linear velocity of an arbitrarily located body-fixed
reference frame. In vector form they are%
\begin{eqnarray}
{^{\mathsf{s}}%
\bm{\Theta}%
^{\mathsf{b}}}\dot{%
\bm{\omega}%
}^{\mathsf{s}}+{\widetilde{%
\bm{\omega}%
}}{^{\mathsf{s}}}%
\hspace{0.2ex}%
{^{\mathsf{s}}%
\bm{\Theta}%
^{\mathsf{b}}}%
\bm{\omega}%
^{\mathsf{s}}+m\widetilde{\mathbf{d}}\ddot{\mathbf{r}} &=&%
\bm{\tau}%
^{\mathsf{s}}  \label{NEhybrid} \\
m(\ddot{\mathbf{r}}+(\dot{\widetilde{%
\bm{\omega}%
}}^{\mathsf{s}}+\widetilde{%
\bm{\omega}%
}^{\mathsf{s}}\widetilde{%
\bm{\omega}%
}^{\mathsf{s}})\mathbf{d}) &=&\mathbf{f}^{\mathsf{h}}.  \notag
\end{eqnarray}%
These are equivalent to the explicit form (\ref{BHBodyFixedExplicit}) of the
body-fixed Newton-Euler equations when expressed in IFR, noting that the
acceleration of the origin of the body-fixed frame in body-fixed
representation is $\dot{\mathbf{v}}^{\mathsf{b}}+\widetilde{%
\bm{\omega}%
}^{\mathsf{b}}\mathbf{v}^{\mathsf{b}}$.

\begin{remark}
The equations (\ref{NEhybrid}) are the Euler-Lagrange equations for the
right-invariant Lagrangian on $SO\left( 3\right) \times {\mathbb{R}}^{3}$.
Using the direct product as rigid body configuration space (in which
rotations and translations are assumed decoupled) has consequences when Lie
group integration schemes are applied to the rigid body equations as \cite%
{BIT2016,MMT2014}, since $SO\left( 3\right) \times {\mathbb{R}}^{3}$ does
not describe frame transformations, i.e. rigid body motions.
\end{remark}

\subsection{A nonholonomically constrained rigid body: Ball rolling on a
horizontal plane%
\label{secBall}%
}

As an example for a constrained rigid body, a solid ball rolling without
slipping on a horizontal plane is considered. It is assumed that the COM is
at the geometric center of the ball. Yet the mass distribution is assumed
inhomogenous so that the ball possess different principle moments of
inertia. This is a simplification of the classical problem of an
inhomogenous rolling ball treated in detail already by Chaplygin in \cite%
{Chaplygin1897a,Chaplygin1897b,Chaplygin1903} and Voronets \cite%
{Woronetz1910}.

\subsubsection{Rolling-without-slipping constraint}

\begin{figure}[t]
\centering\includegraphics[width=0.6
\linewidth]{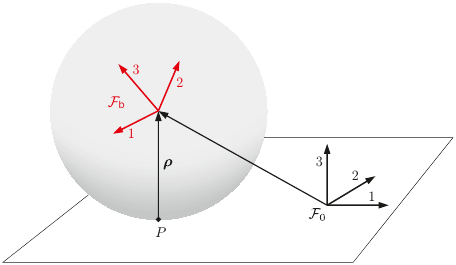}
\caption{Solid ball rolling on a horizontal plane.}
\label{figBall}
\end{figure}
A solid ball with radius $\rho $ rolling on a plane without slipping (Fig. %
\ref{figBall}) is a nonholonomically constrained system. The body-fixed
frame $\mathcal{F}_{\mathsf{b}}$ is attached at the COM of the ball. Denote
with $\mathbf{r}=\left( r_{1},r_{2},r_{3}\right) $ the position vector of
its COM represented in the IFR $\mathcal{F}_{0}$, and with $%
\bm{\varphi}%
=\left( \varphi ^{1},\varphi ^{2},\varphi ^{3}\right) $ the vector of three
parameters describing the orientation. The $n=6$ coordinates describing the
pose of the ball are thus $\mathbf{q}=\left( q^{1},\ldots ,q^{6}\right)
=\left( 
\bm{\varphi}%
,\mathbf{r}\right) $. The hybrid twist $\mathbf{V}=\left( 
\bm{\omega}%
^{\mathsf{s}},\dot{\mathbf{r}}\right) $ is used to describe the velocity of
the ball, where the spatial angular velocity is determined by (\ref{ws1}).
The superscript \textsf{s} is omitted in the following. Denote with $\mathbf{%
n}$ the unit normal vector to the plane resolved in IFR. Then the vector
from the contact point $P$ to the COM is $%
\bm{\rho}%
=\rho \mathbf{n}$, and the rolling-without-slipping condition is%
\begin{equation}
\mathbf{0}=\dot{\mathbf{r}}+\widetilde{%
\bm{\rho}%
}%
\bm{\omega}%
=\dot{\mathbf{r}}+\widetilde{%
\bm{\rho}%
}%
\bm{\Omega}%
\dot{%
\bm{\varphi}%
}  \label{BallConstr}
\end{equation}%
which yields a system of $\bar{m}=3$ Pfaffian constraints (\ref%
{PfaffianConstr}) on $\dot{\mathbf{q}}$, with%
\begin{equation}
u^{\alpha }:=\dot{r}_{\alpha }+\varepsilon _{\alpha jk}\rho _{j}\Omega _{kl}%
\dot{\varphi}^{l},\ \ \alpha =1,2,3.  \label{BallConstr2}
\end{equation}%
The components of the angular velocity constitute $n-\bar{m}=3$ independent
quasi-velocities 
\begin{equation}
u^{I}:=\omega _{\bar{I}},I=4,5,6.
\end{equation}%
Where the notation $\bar{I}:=I-3$ will be used. The constraints (\ref%
{BallConstr2}) along with $\omega _{\bar{I}}=\Omega _{\bar{I}i}\dot{\varphi}%
^{i}$ give rise to the Pfaffian system (\ref{PfaffForm}) with matrix%
\begin{equation}
\mathbf{A}=\left( 
\begin{array}{cc}
\widetilde{%
\bm{\rho}%
}%
\bm{\Omega}
& \mathbf{I} \\ 
\bm{\Omega}
& \mathbf{0}%
\end{array}%
\right) .
\end{equation}%
The solution (\ref{u3}) determines the $\dot{q}^{a}$ in terms of the
Pfaffian forms $u^{a}$, where the inverse of $\mathbf{A}$ is%
\begin{equation}
\mathbf{B}=\left( 
\begin{array}{cc}
\mathbf{0} & 
\bm{\Psi}
\\ 
\mathbf{I} & -\widetilde{%
\bm{\rho}%
}%
\end{array}%
\right) .
\end{equation}%
In components this is%
\begin{eqnarray}
\dot{r}_{\alpha } &=&u^{\alpha }-\varepsilon _{\alpha jk}\rho _{j}\Omega
_{kl}\dot{\varphi}^{l},\ \ \alpha =1,2,3  \label{radot} \\
\dot{\varphi}^{i} &=&\Psi _{\bar{I}}^{i}u^{I},\ \ I=4,5,6.
\end{eqnarray}

\subsubsection{Hamel coefficients}

The Hamel equations (\ref{BHconstr}) involve the coefficients $\gamma
_{IJ}^{a},a=1,\ldots ,6;I,J=4,5,6$. First consider the case $a=K=4,5,6$.
Since $\mathbf{A}$ does not depend on $\mathbf{r}$, it is $\frac{\partial
A_{ab}}{\partial q^{c}}\equiv 0$ for $b,c=4,5,6$, and (\ref{gammaNH}) yields%
\begin{equation}
\gamma _{IJ}^{K}=\left( \frac{\partial \Omega _{\bar{K}i}}{\partial \varphi
^{j}}-\frac{\partial \Omega _{\bar{K}j}}{\partial \varphi ^{i}}\right) \Psi
_{\bar{I}}^{i}\Psi _{\bar{J}}^{j}.  \label{gammaBallTmp}
\end{equation}%
In analogy to (\ref{IntCond2}), using (\ref{ws2}), the expression (\ref%
{gammaBallTmp}) is the definition of the Hamel coefficients (\ref{gammaWs})
for the spatial angular velocity, i.e.%
\begin{equation}
\gamma _{IJ}^{K}=-\varepsilon _{\bar{K}\bar{I}\bar{J}},\ \ I,J,K=4,5,6.
\label{gammaBall1}
\end{equation}%
Noting that $\rho _{i}=\mathrm{const.}$ in (\ref{BallConstr2}), the Hamel
coefficients for $a=\alpha =1,2,3$ are%
\begin{eqnarray}
\gamma _{IJ}^{\alpha } &=&\varepsilon _{\alpha kl}\rho _{k}\left( \frac{%
\partial \Omega _{li}}{\partial \varphi ^{j}}-\frac{\partial \Omega _{lj}}{%
\partial \varphi ^{i}}\right) \Psi _{\bar{I}}^{i}\Psi _{\bar{J}}^{j} \\
&=&-\varepsilon _{\alpha kl}\rho _{k}\varepsilon _{l\bar{I}\bar{J}},\ \ \
\alpha =1,2,3;I,J=4,5,6.  \label{gammaBall2}
\end{eqnarray}

\subsubsection{Hamel equations}

With the assumption that the COM is at the center of the ball, and that it
is rolling on a horizontal plane, its motion is not affected by gravity. The
Lagrangian thus consists of the kinetic energy only. The kinetic energy of
the unconstrained ball, in terms of the velocity $\dot{\mathbf{r}}$ of its
COM and the angular velocity $%
\bm{\omega}%
$, is%
\begin{equation}
T\left( q^{a},\dot{q}^{a}\right) =\frac{1}{2}\omega _{i}\left( \dot{\varphi}%
\right) {^{\mathsf{s}}\Theta _{ij}^{\mathsf{c}}}\left( \varphi \right)
\omega _{j}\left( \dot{\varphi}\right) +\frac{m}{2}\dot{r}_{i}\dot{r}_{i}
\end{equation}%
where $m$ is its mass and ${^{\mathsf{s}}%
\bm{\Theta}%
^{\mathsf{c}}=\mathbf{R}%
\bm{\Theta}%
^{\mathsf{c}}}\mathbf{R}^{T}$ the inertia tensor w.r.t. its COM, as in (\ref%
{EkinHybrid}) with $\mathbf{d}=\mathbf{0}$, resolved in the IFR. Replacing $%
\dot{q}^{a}$ with (\ref{radot}) yields the kinetic energy in terms of $u^{a}$%
\begin{equation}
\mathcal{T}\left( q^{a},u^{a}\right) =\frac{1}{2}u^{I}({^{\mathsf{s}}\Theta
_{\bar{I}\bar{J}}^{\mathsf{c}}}-m\varepsilon _{\bar{I}i\alpha }\varepsilon
_{\alpha j\bar{J}}\rho _{i}\rho _{j})u^{J}+\frac{m}{2}u^{\alpha }u^{\alpha
}-mu^{\alpha }\varepsilon _{\alpha i\bar{I}}\rho _{i}u^{I}.
\end{equation}%
With the Hamel coefficients (\ref{gammaBall1}) and (\ref{gammaBall2}), the
Hamel equations (\ref{BHconstr}) in terms of $u^{a}$ are%
\begin{equation}
\frac{d}{dt}\frac{\partial \mathcal{T}}{\partial u^{I}}-\frac{\partial 
\mathcal{T}}{\partial \varphi ^{i}}\Psi _{\bar{I}}^{i}-\frac{\partial 
\mathcal{T}}{\partial u^{\alpha }}\varepsilon _{\alpha kl}\rho
_{k}\varepsilon _{l\bar{I}\bar{J}}u^{J}-\frac{\partial \mathcal{T}}{\partial
u^{K}}\varepsilon _{\bar{K}\bar{I}\bar{J}}u^{J}=0,\ \ I=4,5,6.
\label{BHBall1}
\end{equation}%
Using $\frac{d}{dt}({^{\mathsf{s}}\Theta _{ij}^{\mathsf{c}}}\omega _{j})={^{%
\mathsf{s}}\Theta _{ij}^{\mathsf{c}}}\dot{\omega}_{j}+\varepsilon
_{ikl}\omega _{k}{^{\mathsf{s}}\Theta _{lj}^{\mathsf{c}}}$, the necessary
partial derivatives are%
\begin{equation}
\frac{d}{dt}\frac{\partial \mathcal{T}}{\partial u^{I}}=({^{\mathsf{s}%
}\Theta _{\bar{I}\bar{J}}^{\mathsf{c}}}-m\varepsilon _{\bar{I}i\alpha
}\varepsilon _{\alpha j\bar{J}}\rho _{i}\rho _{j})\dot{u}^{J}+\varepsilon _{%
\bar{I}\bar{J}i}{^{\mathsf{s}}\Theta _{i\bar{K}}^{\mathsf{c}}}u^{K},\ \frac{%
\partial \mathcal{T}}{\partial u^{\alpha }}=-m\varepsilon _{\alpha i\bar{I}%
}\rho _{i}u^{I}+mu^{\alpha },\frac{\partial \mathcal{T}}{\partial \varphi
^{i}}\Psi _{\bar{I}}^{i}=-\varepsilon _{\bar{I}\bar{J}i}u^{J}{^{\mathsf{s}%
}\Theta _{i\bar{K}}^{\mathsf{c}}}u^{K}.
\end{equation}%
Inserted in (\ref{BHBall1}), and setting $u^{\alpha }=0$, leads to%
\begin{equation}
({^{\mathsf{s}}\Theta _{\bar{I}\bar{J}}^{\mathsf{c}}}-m\varepsilon _{\bar{I}%
i\alpha }\varepsilon _{\alpha j\bar{J}}\rho _{i}\rho _{j})\dot{u}%
^{J}+\varepsilon _{\bar{I}\bar{J}i}u^{J}{^{\mathsf{s}}\Theta _{i\bar{K}}^{%
\mathsf{c}}}u^{K}+m\left( \varepsilon _{\bar{K}i\alpha }\varepsilon _{\alpha
j\bar{L}}\rho _{i}\rho _{r}\varepsilon _{\bar{K}\bar{I}\bar{J}%
}-mu^{K}\varepsilon _{\bar{K}i\alpha }\varepsilon _{\alpha rs}\rho _{i}\rho
_{r}\varepsilon _{s\bar{I}\bar{J}}\right) u^{J}=0,\ \ I=4,5,6.
\label{BHBall2}
\end{equation}%
The last term in (\ref{BHBall2}) vanishes, and replacing $u^{I}=\omega _{%
\bar{I}}$, the Hamel equations of the rolling ball are%
\begin{equation}
{^{\mathsf{s}}\dot{L}_{i}^{\mathsf{P}}}={^{\mathsf{s}}\Theta _{ij}^{\mathsf{P%
}}}\dot{\omega}_{j}^{\mathsf{s}}+\varepsilon _{ijk}\omega _{j}^{\mathsf{s}}{%
^{\mathsf{s}}\Theta _{kl}^{\mathsf{c}}}\omega _{l}^{\mathsf{s}}=0,
\label{BHBall3}
\end{equation}%
where ${^{\mathsf{s}}\Theta _{ij}^{\mathsf{P}}}:={^{\mathsf{s}}\Theta _{ij}^{%
\mathsf{c}}}-m\varepsilon _{irk}\varepsilon _{ksj}\rho _{r}\rho _{s}$ is the
inertia tensor w.r.t. the contact point $P$ expressed in the IFR, and ${^{%
\mathsf{s}}L_{i}^{\mathsf{P}}}:={^{\mathsf{s}}\Theta _{ij}^{\mathsf{P}}}%
\omega _{j}$ is the angular momentum w.r.t. to $P$. Relation (\ref{BHBall3})
is nothing but the condition for conservation of angular momentum. However,
it is important to note that this is only true since the vector $%
\bm{\rho}%
$ is constant. This does not hold if the COM is not at the geometric center.
The vector form of (\ref{BHBall3}) is, with ${^{\mathsf{s}}%
\bm{\Theta}%
^{\mathsf{P}}}:={^{\mathsf{s}}%
\bm{\Theta}%
^{\mathsf{c}}}-m\widetilde{%
\bm{\rho}%
}\widetilde{%
\bm{\rho}%
}$ and ${^{\mathsf{s}}\mathbf{L}^{\mathsf{P}}}:={^{\mathsf{s}}%
\bm{\Theta}%
^{\mathsf{P}}}%
\bm{\omega}%
^{\mathsf{s}}$,%
\begin{equation}
{^{\mathsf{s}}\dot{\mathbf{L}}^{\mathsf{P}}}={^{\mathsf{s}}%
\bm{\Theta}%
^{\mathsf{P}}}\dot{%
\bm{\omega}%
}^{\mathsf{s}}+{\widetilde{%
\bm{\omega}%
}}{^{\mathsf{s}}}%
\hspace{0.2ex}%
{^{\mathsf{s}}%
\bm{\Theta}%
^{\mathsf{c}}}%
\bm{\omega}%
^{\mathsf{s}}=\mathbf{0}.
\end{equation}

\section{Conclusion}

The classical problem of rigid body dynamics in terms of quasi-velocities
has been addressed using the BH-equations and the original definition of the
Hamel coefficients. The Newton-Euler equations are recovered and shown to be
the identical to the Euler-Poincar\'{e} equations, if the body-fixed twists
are used as quasi-velocities, and to the reduced Euler-Lagrange equations,
if the spatial and hybrid representation of rigid body twists are used. In
these cases the Hamel coefficients are identical to the structure constants
of $SE\left( 3\right) $ and $SO\left( 3\right) \times {\mathbb{R}}^{3}$,
respectively. For illustration purpose the BH-equations of a rolling
homogenous ball are derived to demonstrated the Hamel-equations for a
constrained rigid body.

The novel contribution of this paper is the explicit derivation of the Hamel
coefficients for the three most relevant representations of rigid body
twists, without resorting to a description in terms of local coordinates,
using their original definition. This shall complement the classical
treatment of rigid body dynamics by means of the BH-equations. The presented
approach offers a direct link between the classical treatment and the modern
Lie group methods for discrete mechanical systems evolving on a Lie group.

Finally it is to be mentioned that Hamel's variational approach to the
dynamics of constrained rigid bodies is not only relevant for optimal
control \cite{MaruskinBloch2007} but has lead to advanced Lie group modeling
and numerical integration schemes for robots and general multibody systems 
\cite%
{BallZenkov2015,Borri2003,BruelsCardona2010,Hairer2006,MUBOScrews2,ParkChung2005,MUBO2014,CND2015,Zenkov2012}%
.

\begin{appendix}%

\section{Rigid Body Motions}

The motion of a rigid body is represented by the motion of a body-fixed
reference frame $\mathcal{F}_{\mathsf{b}}$, generally located at the body.
The pose of the rigid body is thus described by the frame transformation
from the body-fixed frame $\mathcal{F}_{\mathsf{b}}$ to an IFR $\mathcal{F}%
_{0}$. This frame transformation, and thus the pose of the body, is
determined by the position vector $\mathbf{r}\in {\mathbb{R}}^{3}$ of the
body-fixed frame measured and resolved in the IFR, along with the rotation
matrix $\mathbf{R}\in SO\left( 3\right) $ transforming coordinates of a
vector resolved in a body-fixed frame to those resolved in the IFR.

\subsection{Rotations --Lie Group $SO\left( 3\right) $}

Let $\mathbf{R}=\left( R_{ij}\right) \in SO\left( 3\right) $ be the rotation
matrix transforming from body-fixed to IFR. An arbitrary parameterization in
terms of three variables $%
\bm{\varphi}%
=\left( \varphi ^{1},\varphi ^{2},\varphi ^{3}\right) $ is assumed, i.e. $%
\mathbf{R}=\mathbf{R}\left( 
\bm{\varphi}%
\right) $. Typical choices are Euler or Cardan angles, or the three
components of the product of rotation axis and rotation angle.

The angular velocity vector in \emph{body-fixed representation}, denoted $%
\bm{\omega}%
^{\mathsf{b}}=\left( \omega _{1}^{\mathsf{b}},\omega _{2}^{\mathsf{b}%
},\omega _{3}^{\mathsf{b}}\right) ^{T}$, is defined via the skew-symmetric
angular velocity tensor $\widetilde{%
\bm{\omega}%
}^{\mathsf{b}}=\mathbf{R}^{T}\dot{\mathbf{R}}$. The latter is related to $%
\bm{\omega}%
^{\mathsf{b}}$ by the correspondence%
\begin{equation}
\mathbf{x}\in {\mathbb{R}}^{3}\leftrightarrow \widetilde{\mathbf{x}}\in
so\left( 3\right)  \label{skew}
\end{equation}%
where the components of $\widetilde{\mathbf{x}}=(x_{ij})\in so\left(
3\right) $ are $x_{ij}=\varepsilon _{ikj}x_{k}$. The matrix (\ref{skew}) is
frequently called the cross product matrix since $\widetilde{\mathbf{x}}%
\mathbf{y}=\mathbf{x}\times \mathbf{y}$. With the parameterization in terms
of $%
\bm{\varphi}%
$, the body-fixed angular velocity tensor and vector are determined as%
\begin{eqnarray}
\widetilde{%
\bm{\omega}%
}^{\mathsf{b}} &=&\mathbf{R}^{T}\dot{\mathbf{R}}=\mathbf{R}^{T}\frac{%
\partial \mathbf{R}}{\partial \varphi ^{a}}\dot{\varphi}^{a}=\widetilde{%
\bm{\Omega}%
}{_{a}^{\mathsf{b}}}\dot{\varphi}^{a},\ \ \widetilde{%
\bm{\Omega}%
}{_{a}^{\mathsf{b}}}:=\mathbf{R}^{T}\frac{\partial \mathbf{R}}{\partial
\varphi ^{a}}  \label{w1} \\
\bm{\omega}%
^{\mathsf{b}} &=&\sum_{a}%
\bm{\Omega}%
_{a}^{\mathsf{b}}\dot{\varphi}^{a}=%
\bm{\Omega}%
^{\mathsf{b}}\dot{%
\bm{\varphi}%
},\ \ 
\bm{\Omega}%
^{\mathsf{b}}=\left( 
\bm{\Omega}%
_{1}^{\mathsf{b}},%
\bm{\Omega}%
_{2}^{\mathsf{b}},%
\bm{\Omega}%
_{3}^{\mathsf{b}}\right)  \label{w}
\end{eqnarray}%
where $%
\bm{\Omega}%
_{a}^{\mathsf{b}}$ is the $a$-th column of $%
\bm{\Omega}%
^{\mathsf{b}}=\left( \Omega _{ia}^{\mathsf{b}}\right) $. Written in
component form, these relations are 
\begin{eqnarray}
\omega _{ij}^{\mathsf{b}} &=&\varepsilon _{ikj}\omega _{k}^{\mathsf{b}%
}=R_{ki}\frac{\partial R_{kj}}{\partial \varphi _{a}}\dot{\varphi}^{a}
\label{wij1} \\
&=&\varepsilon _{ikj}\Omega _{ka}^{\mathsf{b}}\dot{\varphi}^{a}  \label{wij2}
\\
\omega _{i}^{\mathsf{b}} &=&\Omega _{ia}^{\mathsf{b}}\dot{\varphi}^{a}
\label{wi}
\end{eqnarray}%
From (\ref{w}) and (\ref{wij1}) follows%
\begin{eqnarray}
\varepsilon _{ikj}\Omega _{ka}^{\mathsf{b}} &=&\frac{\partial \omega _{ij}^{%
\mathsf{b}}}{\partial \dot{\varphi}^{a}}  \label{Aij} \\
\bm{\Omega}%
_{a}^{\mathsf{b}} &=&\frac{\partial 
\bm{\omega}%
^{\mathsf{b}}}{\partial \dot{\varphi}^{a}},\ \Omega _{ia}^{\mathsf{b}}=\frac{%
\partial \omega _{i}^{\mathsf{b}}}{\partial \dot{\varphi}^{a}}  \label{Aa}
\end{eqnarray}%
The inverse relation to (\ref{wi}) is%
\begin{equation}
\dot{\varphi}^{a}={\Psi }{_{i}^{a,\mathsf{b}}}\omega _{i}^{\mathsf{b}}\ \ \ 
\text{with \ \ \ }%
\bm{\Psi}%
^{\mathsf{b}}:=(%
\bm{\Omega}%
^{\mathsf{b}})^{-1}.  \label{Psi}
\end{equation}%
Combining this with (\ref{wij1}) yields%
\begin{equation}
R_{ki}\omega _{ij}^{\mathsf{b}}=\frac{\partial R_{kj}}{\partial \varphi ^{a}}%
{\Psi }{_{r}^{a,\mathsf{b}}}\omega _{r}^{\mathsf{b}},\text{\ \ }%
R_{kr}\varepsilon _{rij}=\frac{\partial R_{kj}}{\partial \varphi ^{a}}{\Psi }%
{_{i}^{a,\mathsf{b}}}.  \label{wIdent}
\end{equation}

In \emph{spatial representation} the angular velocity $%
\bm{\omega}%
^{\mathsf{s}}=\left( \omega _{1}^{\mathsf{s}},\omega _{2}^{\mathsf{s}%
},\omega _{3}^{\mathsf{s}}\right) ^{T}$ is defined by $\widetilde{%
\bm{\omega}%
}^{\mathsf{s}}=\dot{\mathbf{R}}\mathbf{R}^{T}$. In analogy to (\ref{w1}),(%
\ref{wi}), it is expressed using the parameterization as%
\begin{eqnarray}
\widetilde{%
\bm{\omega}%
}^{\mathsf{s}} &=&\dot{\mathbf{R}}\mathbf{R}^{T}=\frac{\partial \mathbf{R}}{%
\partial \varphi ^{a}}\mathbf{R}^{T}\dot{\varphi}^{a}=\widetilde{%
\bm{\Omega}%
}{_{a}^{\mathsf{s}}}\dot{\varphi}^{a},\ \ \widetilde{%
\bm{\Omega}%
}{_{a}^{\mathsf{s}}}:=\frac{\partial \mathbf{R}}{\partial \varphi ^{a}}%
\mathbf{R}^{T}  \label{ws1} \\
\bm{\omega}%
^{\mathsf{s}} &=&\sum_{a}%
\bm{\Omega}%
_{a}^{\mathsf{s}}\dot{\varphi}^{a}=%
\bm{\Omega}%
^{\mathsf{s}}\dot{%
\bm{\varphi}%
},\ \ 
\bm{\Omega}%
^{\mathsf{s}}=\left( 
\bm{\Omega}%
_{1}^{\mathsf{s}},%
\bm{\Omega}%
_{2}^{\mathsf{s}},%
\bm{\Omega}%
_{3}^{\mathsf{s}}\right) .  \label{ws2}
\end{eqnarray}%
The inverse of this relation is $\dot{\varphi}^{a}={\Psi }{_{i}^{a,\mathsf{s}%
}}\omega _{i}^{\mathsf{s}}$, with $%
\bm{\Psi}%
^{\mathsf{s}}:=(%
\bm{\Omega}%
^{\mathsf{s}})^{-1}$.

With the isomorphism (\ref{skew}), the Lie bracket on $so\left( 3\right) $
can be expressed in matrix or vector representation, respectively,%
\begin{equation}
\lbrack \widetilde{\mathbf{x}},\widetilde{\mathbf{y}}]=\widetilde{\mathbf{x}}%
\widetilde{\mathbf{y}}-\widetilde{\mathbf{y}}\widetilde{\mathbf{x}},\ \ [%
\mathbf{x},\mathbf{y}]=\widetilde{\mathbf{x}}\mathbf{y}=\mathbf{ad}_{\mathbf{%
x}}\mathbf{y}  \label{SO3Lie}
\end{equation}%
with $\mathbf{x},\mathbf{y}\in {\mathbb{R}}^{3}$, where $\mathbf{ad}_{%
\mathbf{x}}=\widetilde{\mathbf{x}}$. The respective components are%
\begin{equation}
\lbrack \widetilde{\mathbf{x}},\widetilde{\mathbf{y}}]_{ij}=\varepsilon
_{ikj}\varepsilon _{krs}x_{r}y_{s},\ \ [\mathbf{x},\mathbf{y}%
]_{i}=\varepsilon _{ijk}x_{j}y_{k}  \label{SO3Lie2}
\end{equation}%
which follows from the identity $\varepsilon _{kij}\varepsilon _{krs}=\delta
_{ir}\delta _{js}-\delta _{is}\delta _{jr}$. That is, the structure
constants of $SO\left( 3\right) $ are given by the Levi-Civita symbol $%
\varepsilon _{ijk}$.

\subsection{Spatial rigid body motions%
\label{secSE3}%
}

\subsubsection{Rigid body motions as frame transformations --Lie Group $%
SE\left( 3\right) $}

The motion of a rigid body is described by the motion of a body-fixed frame $%
\mathcal{F}$. The configuration (pose, posture) of the body is described by
a frame transformation from body-fixed to inertial frame. This is described
by the rotation matrix $\mathbf{R}\in SO\left( 3\right) $, transforming
coordinates of a vector when expressed in the body-fixed frame to those when
expressed in inertia frame, and the position vector $\mathbf{r}\in {\mathbb{R%
}}^{3}$ of the origin of the body-fixed frame measured in the IFR. Frame
transformations form the Lie group $SE\left( 3\right) $, and are denoted as $%
C=\left( \mathbf{R},\mathbf{r}\right) \in SE\left( 3\right) $, and
frequently represented by 'homogenous transformation' matrices%
\begin{equation}
\mathbf{C}=\left( 
\begin{array}{cc}
\mathbf{R} & \mathbf{r} \\ 
\mathbf{0} & 1%
\end{array}%
\right) \in SE\left( 3\right)  \label{C}
\end{equation}%
that transform point coordinates measured in a body-fixed frame to those
measured in the IFR. For details on $SE\left( 3\right) $ the reader is
referred to \cite{Selig}.

The transition from one configuration to another is described by a repeated
frame transformation, and achieved by the group multiplication $C_{1}\cdot
C_{2}=\left( \mathbf{R}_{1}\mathbf{R}_{2},\mathbf{r}_{1}+\mathbf{R}_{1}%
\mathbf{r}_{2}\right) $, or in matrix form $\mathbf{C}_{1}\mathbf{C}_{2}$.
This multiplication reveals that $SE\left( 3\right) =SO\left( 3\right)
\ltimes {\mathbb{R}}^{3}$ is the semi-direct product of the rotation group
and the translation group.

The \emph{body-fixed twist} of the rigid body is defined as the
left-invariant vector field $T_{C}L_{C^{-1}}\dot{C}=(\widetilde{%
\bm{\omega}%
}^{\mathsf{b}},\mathbf{v}^{\mathsf{b}})$. This is usually expressed in
matrix form as%
\begin{equation}
\mathbf{C}^{-1}\dot{\mathbf{C}}=\left( 
\begin{array}{cc}
\widetilde{%
\bm{\omega}%
}^{\mathsf{b}} & \mathbf{v}^{\mathsf{b}} \\ 
\mathbf{0} & 0%
\end{array}%
\right) =\widehat{\mathbf{V}}^{\mathsf{b}}\in se\left( 3\right)
\label{BodyFixedTwist}
\end{equation}%
with the body-fixed angular velocity $%
\bm{\omega}%
^{\mathsf{b}}\in {\mathbb{R}}^{3}$ and translational velocity $\mathbf{v}^{%
\mathsf{b}}=\mathbf{R}^{T}\dot{\mathbf{r}}\in {\mathbb{R}}^{3}$. This
defines $\mathbf{V}^{\mathsf{b}}\in {\mathbb{R}}^{6}$ according to%
\begin{equation}
\mathbf{V}^{\mathsf{b}}=\left( 
\begin{array}{c}
\bm{\omega}%
^{\mathsf{b}} \\ 
\mathbf{v}^{\mathsf{b}}%
\end{array}%
\right) \in {\mathbb{R}}^{6}\leftrightarrow \widehat{\mathbf{V}}^{\mathsf{b}%
}=\left( 
\begin{array}{cc}
\widetilde{%
\bm{\omega}%
}^{\mathsf{b}} & \mathbf{v}^{\mathsf{b}} \\ 
\mathbf{0} & 0%
\end{array}%
\right) \in se\left( 3\right) .  \label{hat}
\end{equation}

The \emph{spatial twist} of the rigid body is defined as the right-invariant
vector fields $T_{C}R_{C^{-1}}\dot{C}=(\widetilde{%
\bm{\omega}%
}^{\mathsf{s}},\mathbf{v}^{\mathsf{s}})$, and in matrix form%
\begin{equation}
\dot{\mathbf{C}}\mathbf{C}^{-1}=\left( 
\begin{array}{cc}
\widetilde{%
\bm{\omega}%
}^{\mathsf{s}} & \mathbf{v}^{\mathsf{s}} \\ 
\mathbf{0} & 0%
\end{array}%
\right) =\widehat{\mathbf{V}}^{\mathsf{s}}\in se\left( 3\right)
\label{SpatialTwist}
\end{equation}%
where $\mathbf{v}^{\mathsf{s}}=\dot{\mathbf{r}}-\widetilde{%
\bm{\omega}%
}^{\mathsf{s}}\mathbf{r}\in {\mathbb{R}}^{3}$ is the velocity of the
body-fixed point momentarily coinciding with the origin of the IFR. This is
also called the fixed pole formulation \cite{Bottasso1998}. The
corresponding spatial twist vector is $\mathbf{V}^{\mathsf{b}}=(%
\bm{\omega}%
^{\mathsf{s}},\mathbf{v}^{\mathsf{s}})$.

It will be necessary to express the relation (\ref{hat}) of matrix $\widehat{%
\mathbf{V}}\in se\left( 3\right) $ and vector $\mathbf{V}\in {\mathbb{R}}%
^{6} $ in components. Notice again that the basis on $se\left( 3\right) $ is
deduced from an orthogonal Cartesian frame. To this end, in the following,
Greek letters run over the range $1,\ldots ,4$, while italic letters run
over $1,\ldots ,6$. The components of $\widehat{\mathbf{V}}=\left( V_{\alpha
\beta }\right) \in se\left( 3\right) $ are $V_{\alpha \beta }=h_{\alpha
i\beta }V_{i},i=1,\ldots ,6,\alpha ,\beta =1,\ldots ,4$, where the constant
coefficients%
\begin{equation}
h_{\alpha i\beta }=\left\{ 
\begin{array}{rl}
\varepsilon _{\alpha i\beta }, & \alpha ,\beta ,i=1,2,3 \\ 
\delta _{\alpha \ibar4}, & \alpha =1,2,3;\beta =4;i=4,5,6 \\ 
0, & \text{otherwise}%
\end{array}%
\right.  \label{h}
\end{equation}%
assign the six components of $\mathbf{V}\in {\mathbb{R}}^{6}$ to the six
non-zero components of $\widehat{\mathbf{V}}\in se\left( 3\right) $,
similarly to the Levi-Civita symbol in (\ref{skew}) and (\ref{wij1}). Here
the notation $\ibar=i-3$ has been used.

The Lie bracket on $se\left( 3\right) $ can be expressed in matrix and
vector representation, respectively, as%
\begin{equation}
\lbrack \widehat{\mathbf{X}},\widehat{\mathbf{Y}}]=\widehat{\mathbf{X}}%
\widehat{\mathbf{Y}}-\widehat{\mathbf{Y}}\widehat{\mathbf{X}}=\left( 
\begin{array}{cc}
\widetilde{%
\bm{\omega}%
}_{X}\widetilde{%
\bm{\omega}%
}_{Y}-\widetilde{%
\bm{\omega}%
}_{Y}\widetilde{%
\bm{\omega}%
}_{X} & \widetilde{%
\bm{\eta}%
}_{X}%
\bm{\omega}%
_{Y}+\widetilde{%
\bm{\omega}%
}_{X}%
\bm{\eta}%
_{Y} \\ 
\mathbf{0} & 0%
\end{array}%
\right) ,\ \ [\mathbf{X},\mathbf{Y}]=\mathbf{ad}_{\mathbf{X}}\mathbf{Y}%
=\left( 
\begin{array}{c}
\widetilde{%
\bm{\omega}%
}_{X}%
\bm{\omega}%
_{Y} \\ 
\widetilde{%
\bm{\omega}%
}_{X}%
\bm{\eta}%
_{Y}+\widetilde{%
\bm{\eta}%
}_{X}%
\bm{\omega}%
_{Y}%
\end{array}%
\right)  \label{SE3Lie}
\end{equation}%
with matrix%
\begin{equation}
\mathbf{ad}_{\mathbf{X}}=\left( 
\begin{array}{cc}
\widetilde{%
\bm{\omega}%
} & \mathbf{0} \\ 
\widetilde{%
\bm{\eta}%
} & \widetilde{%
\bm{\omega}%
}%
\end{array}%
\right) .  \label{adSE3}
\end{equation}%
The correspondence (\ref{hat}), i.e. the isomorphism $\widehat{[\mathbf{X},%
\mathbf{Y}]}=[\widehat{\mathbf{X}},\widehat{\mathbf{Y}}]$, yields the
components of the Lie bracket in matrix representation and vector
representation of $se\left( 3\right) $, respectively, as%
\begin{equation}
\lbrack \widehat{\mathbf{X}},\widehat{\mathbf{Y}}]_{\alpha \beta }=h_{\alpha
i\beta }c_{ijk}X_{j}Y_{k},\ \ [\mathbf{X},\mathbf{Y}]_{i}=c_{ijk}X_{j}Y_{k}
\label{SE3Lie2}
\end{equation}%
where $c_{ijk}$ are the structure constants of $se\left( 3\right) $,%
\begin{equation}
c_{ijk}=\left\{ 
\begin{array}{rl}
\varepsilon _{ijk}, & \alpha ,\beta ,i=1,2,3 \\ 
\varepsilon _{\ibar j\bar{k}}, & i,k=4,5,6;j=1,2,3 \\ 
\varepsilon _{\ibar\bar{j}k}, & i,j=4,5,6;k=1,2,3 \\ 
0, & \text{otherwise.}%
\end{array}%
\right.  \label{cijk}
\end{equation}%
It is apparent from the structure constants that $SE\left( 3\right) $ is the
semi-direct product. The correspondence (\ref{SE3Lie2}) of the components is
established by the identity%
\begin{equation}
h_{\alpha i\rho }h_{\rho j\beta }-h_{\alpha j\rho }h_{\rho i\beta
}=h_{\alpha k\beta }c_{kij}.  \label{hc}
\end{equation}%
The rigid body transformation is parameterized in terms of canonical
coordinates $\mathbf{q}=\left( q^{1},\ldots ,q^{6}\right) $ on $SE\left(
3\right) $, e.g. screw coordinates or three angles together with the three
components of the position vector. The body-fixed twist is then expressed in
terms of their time derivatives as%
\begin{eqnarray}
\widehat{\mathbf{V}}^{\mathsf{b}} &=&\mathbf{C}^{-1}\dot{\mathbf{C}}=\sum_{a}%
\mathbf{C}^{-1}\frac{\partial \mathbf{C}}{\partial q^{a}}\dot{q}^{a}=\sum_{a}%
\widehat{\mathbf{J}}_{a}\dot{q}^{a},\ \ \widehat{\mathbf{J}}_{a}:=\mathbf{C}%
^{-1}\frac{\partial \mathbf{C}}{\partial q^{a}}  \label{V1} \\
\mathbf{V}^{\mathsf{b}} &=&\sum_{a}\mathbf{J}_{a}\dot{q}^{a}=\mathbf{J}\dot{%
\mathbf{q}},\ \ \mathbf{J}=\left( \mathbf{J}_{1},\ldots ,\mathbf{J}%
_{6}\right)  \label{V2}
\end{eqnarray}%
where where $\mathbf{J}_{a}$ is the $a$-th column of $\mathbf{J}=\left(
J_{ia}\right) $. Denote the inverse of the transformation matrix (\ref{C})
with $\mathbf{D:=C}^{-1}$. Then, with $\mathbf{C}=\left( C_{\alpha \beta
}\right) $, $\mathbf{D}=\left( D_{\alpha \beta }\right) $ and $\mathbf{V}%
=\left( V_{i}\right) $, $\widehat{\mathbf{V}}=\left( V_{\alpha \beta
}\right) $, the relations (\ref{V1}) and (\ref{V2}) are in component form%
\begin{eqnarray}
V_{i} &=&J_{ia}\dot{q}^{a}  \label{Va} \\
V_{\alpha \beta } &=&h_{\alpha i\beta }V_{i}=h_{\alpha i\beta }J_{ia}\dot{q}%
^{a}=D_{\alpha \sigma }\frac{\partial C_{\sigma \beta }}{\partial q^{a}}\dot{%
q}^{a}=-\frac{\partial D_{\alpha \sigma }}{\partial q^{a}}C_{\sigma \beta }%
\dot{q}^{a}.  \label{Vab3}
\end{eqnarray}%
These relations immediately give rise to the following expression for the
components of $\mathbf{J}=\left( J_{ia}\right) $%
\begin{eqnarray}
\mathbf{J}_{a} &=&\frac{\partial \mathbf{V}}{\partial \dot{q}^{a}},\ J_{ia}=%
\frac{\partial V_{i}}{\partial \dot{q}^{a}}  \label{Vab1} \\
h_{\alpha i\beta }J_{ia} &=&\frac{\partial V_{\alpha \beta }}{\partial \dot{q%
}^{a}}.  \label{Vab2}
\end{eqnarray}%
The inverse relation to (\ref{Va}) is 
\begin{equation}
\dot{q}^{a}=G_{i}^{a}V_{i}\text{ \ \ \ with \ \ \ }\mathbf{G=J}^{-1}.
\label{Gai}
\end{equation}%
Inserting (\ref{Gai}) in (\ref{Vab3}) yields 
\begin{equation}
\frac{\partial C_{\sigma \beta }}{\partial q^{a}}G_{i}^{a}=C_{\sigma \alpha
}h_{\alpha i\beta },\ \ \ \frac{\partial D_{\alpha \sigma }}{\partial q^{a}}%
G_{i}^{a}=-h_{\alpha i\beta }D_{\beta \sigma }.  \label{wIdentV}
\end{equation}

\subsubsection{Rigid body motions as decoupled rotations and translations
--Lie Group $SO\left( 3\right) \times {\mathbb{R}}^{3}$}

It is a fundamental property of rigid body motions, which is clear from the
frame transformation (\ref{C}) of point coordinates, that rotations and
translations are coupled, but independent. Nevertheless, in many
formulations in computational dynamics and multibody system dynamics,
translations and rotations are treated decoupled \cite%
{ShabanaBook2013,BruelsCardona2010}. This is motivated by using the NE
equations relative to the COM (i.e. $\mathbf{d}=\mathbf{0}$ in (\ref%
{NEhybrid})). The underlying geometric model is the direct product $SO\left(
3\right) \times {\mathbb{R}}^{3}$ with group operation $C_{1}\cdot
C_{2}=\left( \mathbf{R}_{1}\mathbf{R}_{2},\mathbf{r}_{1}+\mathbf{r}%
_{2}\right) $. This does clearly not describe general rigid body motions.

Let $%
\bm{\omega}%
^{\mathsf{s}}$ be the spatial angular velocity, and $\dot{\mathbf{r}}$ be
the velocity of the origin of a body-fixed frame represented in the IFR.
Then, 
\begin{equation}
\mathbf{V}^{\mathrm{h}}=\left( 
\bm{\omega}%
^{\mathsf{s}},\dot{\mathbf{r}}\right)  \label{Vh}
\end{equation}%
is called the \emph{hybrid form} of the body twist since it is the twist
measured at the origin of the body-fixed frame but resolved in the IFR \cite%
{Murray}. The hybrid twist can be defined as the right-invariant vector
field $T_{C}R_{C^{-1}}\dot{C}=(\widetilde{%
\bm{\omega}%
}^{\mathsf{s}},\dot{\mathbf{r}})$, for $C\in SO\left( 3\right) \times {%
\mathbb{R}}^{3}$.

\newpage

\section{List of Symbols%
\label{secSymbols}%
}

\begin{tabular}{lrl}
$\mathcal{F}_{0}$ & -- & spatial inertial frame (IFR) \\ 
$\mathcal{F}_{\mathsf{b}}$ & -- & body-fixed reference frame at arbitrary
location \\ 
$\mathcal{F}_{\mathsf{c}}$ & -- & body-fixed reference frame at COM \\ 
$\mathbf{R}=\left( R_{ij}\right) \in SO\left( 3\right) $ & -- & rotation
matrix transforming coordinates of a vector resolved in body frame \\ 
&  & to those when resolved in spatial inertial frame \\ 
$\mathbf{r}=\left( r_{i}\right) \in {\mathbb{R}}^{3}$ & -- & position vector
of the origin of the body-fixed frame expressed in IFR \\ 
$\mathbf{C}=\left( C_{\alpha \beta }\right) \in SE\left( 3\right) $ & -- & 
homogenous transformation matrix from body-fixed to inertial frame \\ 
$%
\bm{\omega}%
^{\mathsf{b}}=\left( \omega _{i}^{\mathsf{b}}\right) \in {\mathbb{R}}^{3}$ & 
-- & coordinate vector of angular of $\mathcal{F}_{\mathsf{b}}$ relative to $%
\mathcal{F}_{0}$ resolved in $\mathcal{F}_{\mathsf{b}}$ (body-fixed
representation) \\ 
$%
\bm{\omega}%
^{\mathsf{s}}=\left( \omega _{i}^{\mathsf{s}}\right) \in {\mathbb{R}}^{3}$ & 
-- & coordinate vector of angular of $\mathcal{F}_{\mathsf{b}}$ relative to $%
\mathcal{F}_{0}$ resolved in $\mathcal{F}_{0}$ (spatial representation) \\ 
$\mathbf{v}^{\mathsf{b}}=\left( v_{i}^{\mathsf{b}}\right) \in {\mathbb{R}}%
^{3}$ & -- & velocity of origin of $\mathcal{F}_{\mathsf{b}}$ measure and
resolved in $\mathcal{F}_{0}$ \\ 
$\mathbf{v}^{\mathsf{s}}=\left( v_{i}^{\mathsf{s}}\right) \in {\mathbb{R}}%
^{3}$ & -- & velocity of the point of a body coinciding with the origin of $%
\mathcal{F}_{0}$ measure and resolved in $\mathcal{F}_{0}$ \\ 
$\widetilde{%
\bm{\omega}%
}=\left( \omega _{ij}\right) \in so\left( 3\right) $ & -- & skew symmetric
matrix associated to vector $%
\bm{\omega}%
\in {\mathbb{R}}^{3}$, where $\omega _{ij}:=\varepsilon _{ikj}\omega _{k}$
\\ 
$\mathbf{V}^{\mathsf{b}}=\left( V_{i}^{\mathsf{b}}\right) \in {\mathbb{R}}%
^{6}$ & -- & twist vector defined as $\mathbf{V}^{\mathsf{b}}=\left( 
\bm{\omega}%
^{\mathsf{b}},\mathbf{v}^{\mathsf{b}}\right) $ in body-fixed representation
\\ 
$\mathbf{V}^{\mathsf{s}}=\left( V_{i}^{\mathsf{s}}\right) \in {\mathbb{R}}%
^{6}$ & -- & twist vector defined as $\mathbf{V}^{\mathsf{s}}=\left( 
\bm{\omega}%
^{\mathsf{s}},\mathbf{v}^{\mathsf{s}}\right) $ in spatial representation \\ 
$\mathbf{V}^{\mathsf{h}}=\left( V_{i}^{\mathsf{h}}\right) \in {\mathbb{R}}%
^{6}$ & -- & twist vector defined as $\mathbf{V}^{\mathsf{h}}=\left( 
\bm{\omega}%
^{\mathsf{s}},\dot{\mathbf{r}}\right) $ in hybrid representation \\ 
$\widehat{\mathbf{V}}=\left( V_{\alpha \beta }\right) \in se\left( 3\right) $
& -- & representation of the twist by a $4\times 4$-matrix (twist tensor),
where $V_{\alpha \beta }:=h_{\alpha k\beta }V_{k}$ with $h_{\alpha k\beta }$
in (\ref{h}) \\ 
$%
\bm{\theta}%
^{\mathsf{s}}=\left( \theta _{i}^{\mathsf{s}}\right) $ & -- & 
quasi-coordinates for the orientation of the body-fixed frame \\ 
$%
\bm{\theta}%
^{\mathsf{b}}=\left( \theta _{i}^{\mathsf{b}}\right) $ &  & resolved in
spatial inertial frame and body-fixed frame, respectively \\ 
$%
\bm{\rho}%
^{\mathsf{s}}=\left( \rho _{i}^{\mathsf{s}}\right) $ & -- & 
quasi-coordinates for the position of the origin of body-fixed frame \\ 
$%
\bm{\rho}%
^{\mathsf{b}}=\left( \rho _{i}^{\mathsf{b}}\right) $ &  & resolved in
spatial inertial frame and body-fixed frame, respectively \\ 
&  & These are summarized as $%
\bm{\pi}%
^{\mathsf{b}}=(%
\bm{\theta}%
^{\mathsf{b}},%
\bm{\rho}%
^{\mathsf{b}})$ and $%
\bm{\pi}%
^{\mathsf{s}}=(%
\bm{\theta}%
^{\mathsf{s}},%
\bm{\rho}%
^{\mathsf{s}})$. \\ 
$%
\bm{\Theta}%
^{\mathsf{b}}=\left( \Theta _{ij}^{\mathsf{b}}\right) $ & -- & Inertia
tensor w.r.t. to the body-fixed frame \\ 
$%
\bm{\Theta}%
^{\mathsf{s}}=\left( \Theta _{ij}^{\mathsf{s}}\right) $ & -- & Inertia
tensor w.r.t. to the body-fixed point coinciding with IFR origin resolved in
IFR \\ 
$^{\mathsf{s}}%
\bm{\Theta}%
^{\mathsf{b}}=\left( ^{\mathsf{s}}\Theta _{ij}^{\mathsf{b}}\right) $ & -- & 
Inertia tensor w.r.t. origin of body-fixed frame resolved in IFR \\ 
$^{\mathsf{s}}%
\bm{\Theta}%
^{\mathsf{c}}=\left( ^{\mathsf{s}}\Theta _{ij}^{\mathsf{c}}\right) $ & -- & 
Inertia tensor w.r.t. COM resolved in IFR \\ 
$q^{a}$ & -- & local coordinates, $a=1,\ldots ,n$ \\ 
$\pi ^{a}$ & -- & quasi-coordinates \\ 
$u^{a}$ & -- & quasi-velocity coordinates \\ 
$%
\bm{\Omega}%
^{\mathsf{b}}=\left( \Omega _{ia}^{\mathsf{b}}\right) $ & -- & coefficients
of the kinematic relation $\omega _{i}^{\mathsf{b}}=\Omega _{ia}^{\mathsf{b}}%
\dot{\varphi}^{a}$ (superscript \textsf{s} for spatial representation) \\ 
$%
\bm{\Psi}%
^{\mathsf{b}}=({\Psi }{_{i}^{a,\mathsf{b}}})$ & -- & coefficients of the
inverse kinematic relation $\dot{\varphi}^{a}={\Psi }{_{i}^{a,\mathsf{b}}}%
\omega _{i}^{\mathsf{b}}$ \\ 
$\mathbf{J}=\left( J_{ia}\right) $ & -- & coefficients of the kinematic
relation $V_{i}=J_{ia}\dot{q}^{a}$ \\ 
$\mathbf{G}=\left( G_{i}^{a}\right) $ & -- & coefficients of the inverse
kinematic relation $\dot{q}^{a}=G_{i}^{a}V_{i}$ \\ 
$n$ & -- & number of (generalized) coordinates \\ 
$\bar{m}$ & -- & number of Pfaffian constraints \\ 
$a,b,c,\ldots $ & -- & indices running over all coordinates: $a=1,\ldots ,n$
\\ 
$I,K,L,\ldots $ & -- & indices of independent velocity coordinates: $I=\bar{m%
}+1,\ldots ,n$ \\ 
$\alpha ,\beta ,\gamma ,\ldots $ & -- & indices of constraints equations and
of dependent velocity coordinates: $\alpha =1,\ldots ,\bar{m}$ \\ 
&  & indices of an $se\left( 3\right) $-tensor: $\alpha =1,\ldots ,4$ \\ 
$\bar{I}$ & -- & shifted index $\bar{I}:=I-3$ \\ 
$\gamma _{ij}^{k}$ & -- & Hamel coefficients of a rigid body \\ 
$\gamma ^{{}}$ & -- & Hamel coefficients of a rigid body \\ 
$SE\left( 3\right) =SO\left( 3\right) \ltimes {\mathbb{R}}^{3}$ & -- & 
special Euclidean group (rigid body motion group) \\ 
$SO\left( 3\right) $ & -- & special orthogonal group (rotation group) \\ 
\textsf{b} & -- & indicates body-fixed representation (these are not indices)
\\ 
\textsf{s} & -- & indicates spatial representation \\ 
\textsf{h} & -- & indicates hybrid representation%
\end{tabular}

\section{List of Abbreviations%
\label{secAbbr}%
}

\begin{tabular}{lrl}
IFR & -- & Inertial Frame \\ 
COM & -- & Center Of Mass \\ 
BH & -- & Boltzmann-Hamel%
\end{tabular}%
\clearpage%
\end{appendix}%

\section*{Acknowledgment}

The author would like Dr. Abhinandan Jain from NASA Jet Propulsion
Laboratory for pointing out the lack of a explicit coordinate-free
derivation of the Hamel coefficients for the rigid body. The author
acknowledges support by the LCM-K2 Center within the framework of the
Austrian COMET-K2 program.

\end{document}